\documentclass{article}

    \PassOptionsToPackage{numbers, compress}{natbib}

\usepackage{xspace}

\usepackage[preprint]{neurips_2024}

\usepackage[utf8]{inputenc} 
\usepackage[T1]{fontenc}    
\usepackage[pagebackref,breaklinks,colorlinks]{hyperref}
\usepackage{url}            
\usepackage{booktabs}       
\usepackage{amsfonts}       
\usepackage{nicefrac}       
\usepackage{microtype}      
\usepackage{xcolor}         

\usepackage{marvosym}
\usepackage{graphicx}
\usepackage{caption}
\usepackage{amsmath}
\usepackage{cleveref}
\usepackage{enumitem}
\usepackage{multirow} 
\usepackage{cleveref}

\newcommand{\eg}{e.g.,}

\newcommand{\methodname}{Make-it-Real\xspace}

\definecolor{pleasantred}{RGB}{218, 98, 125}  
\definecolor{pleasantgreen}{RGB}{117, 189, 66}  
\definecolor{pleasantblue}{RGB}{83, 158, 199}  
\definecolor{pleasantgray}{RGB}{175, 171, 171}

\title{
Make-it-Real:
Unleashing Large Multimodal Model for Painting 3D Objects with Realistic Materials}

\author{
Ye Fang$^{1,4}$\thanks{Equal Contribution} \;, Zeyi Sun$^{2,4*}$, Tong Wu$^{3,6} ${\textsuperscript{\Letter}}, Jiaqi Wang$^{4}$, \\ \textbf{Ziwei Liu}$^{5}$, \textbf{Gordon Wetzstein}$^{6}$, \textbf{Dahua Lin}$^{3,4} ${\textsuperscript{\Letter}}\\
$^1$Fudan University \quad
$^2$Shanghai Jiao Tong University \quad 
$^3$The Chinese University of Hong Kong \quad  \\
$^4$Shanghai AI Laboratory \quad $^5$S-Lab, Nanyang Technological University \quad $^6$Stanford University\\
\vspace{-10mm}
}

\begin{document}

\maketitle
\begin{figure}[htbp]
    \centering
    \vspace{0pt}
    \includegraphics[width=\textwidth]{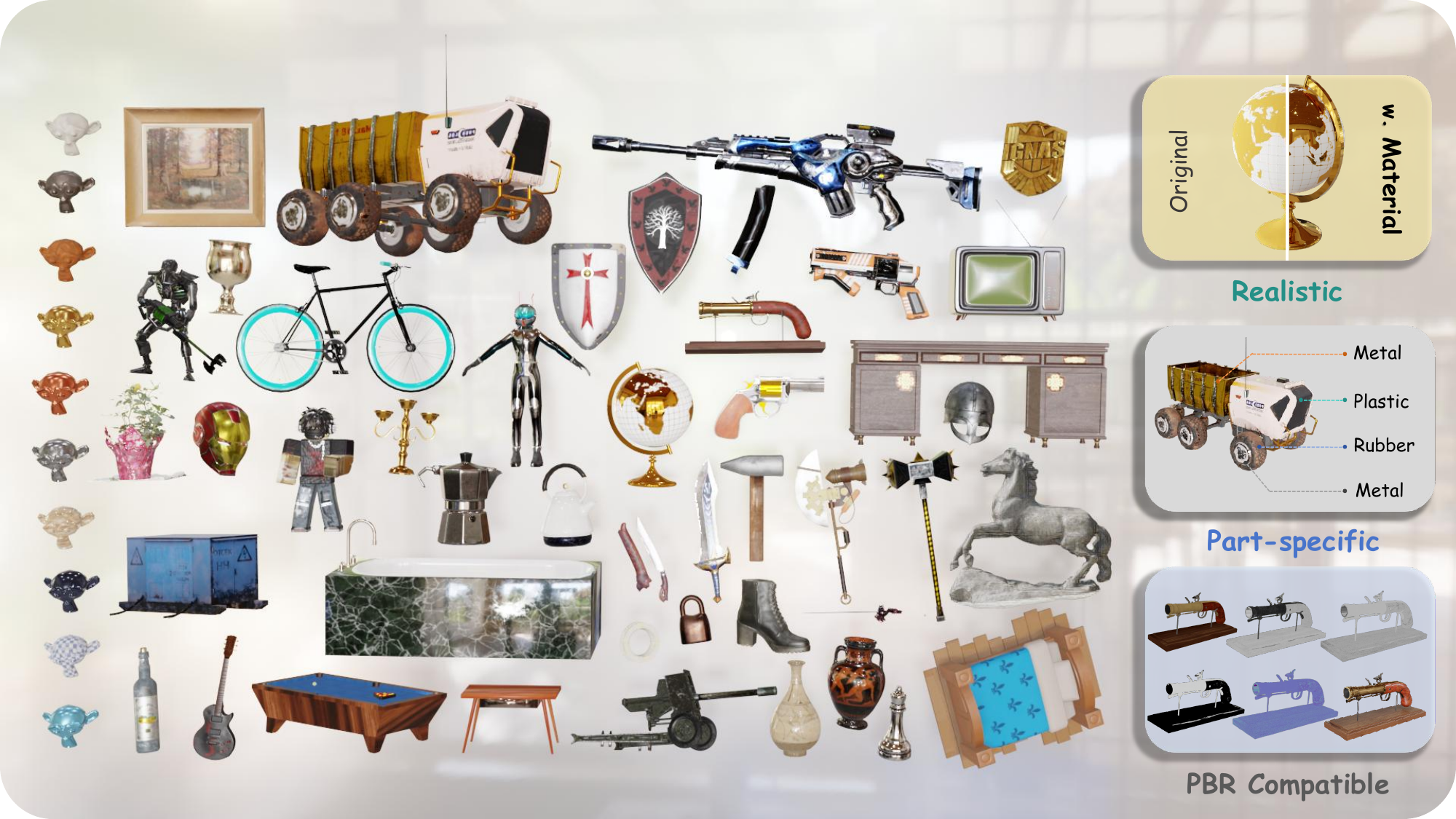}
    \vspace{-10pt}
    \caption{\textbf{Usage of Make-it-Real}. Our method can refine a wide range of albedo-map-only 3D objects from both CAD design and generative models. Our method enhances the realism of objects, enables part-specific material assignment to objects and generate PBR maps that are compatible with downstream engines.}
    \label{fig:teaser}
    \vspace{-15pt}
\end{figure}

\begin{abstract}
Physically realistic materials are pivotal in augmenting the realism of 3D assets across various applications and lighting conditions. However, existing 3D assets and generative models often lack authentic material properties. Manual assignment of materials using graphic software is a tedious and time-consuming task. In this paper, we exploit advancements in Multimodal Large Language Models (MLLMs), particularly GPT-4V, to present a novel approach, \textbf{Make-it-Real}: \textbf{1)} We demonstrate that GPT-4V can effectively recognize and describe materials. \textbf{2)} Utilizing a combination of visual cues and hierarchical text prompts, GPT-4V precisely identifies and aligns materials with the corresponding components of 3D objects. \textbf{3)} The correctly matched materials are then meticulously applied as reference for the new SVBRDF material generation according to the original albedo map, significantly enhancing their visual authenticity. Make-it-Real offers a streamlined integration into the 3D content creation workflow, showcasing its utility as an essential tool for developers of 3D assets. Our project website is at: \url{https://SunzeY.github.io/Make-it-Real/}.

\end{abstract}
\section{Introduction}
\label{sec:intro}

High-quality materials are important for the nuanced inclusion of view-dependent and lighting-dependent effects in 3D assets for traditional graphics pipelines, critical for achieving realism in gaming, online product showcasing, and virtual/augmented reality.
However, many existing assets and generated objects often lack realistic material properties, limiting their application in downstream tasks. Furthermore, creating hand-designed realistic textures necessitates specialized graphic software and involves a laborious and time-consuming process, compounded by significant creative challenges~\cite{labschutz2011content}.

Traditional computer graphics methods have been either manually creating materials or reconstructing them from physical measurements. Emerging text-to-3D generative models~\cite{Chen2023Fantasia3DDG, instant3d2023, Chan2021EfficientG3, chen2022gdna, dong2023ag3d, richardson2023texture} and image-to-3D generative models~\cite{xu2024instantmesh,tochilkin2024triposr,hong2023lrm,wang2023pflrm,openlrm,wang2023imagedream} are successful in creating complex geometries and detailed appearances, but they struggle to generate physically realistic materials, hampering their practical applicability. 
Recent studies have also explored advanced aspects of appearance generation~\cite{chen2022tango,youwang2023paint,xu2023matlaber,chen2023fantasia3d, lopes2023material}. However, they often rely on simplified material models. For instance, ~\cite{lopes2023material} lacks the ability to produce metallic maps. All these approaches do not generate corresponding displacement and height maps, which restricts the diversity and realism of the generated materials, especially regarding depth and tactile qualities. Furthermore, these methods typically require relatively long training and inference time. Considering the abundance of high-quality 3D assets online~\cite{objaverse, objaverseXL} that lack material attributes, and the maturity of 3D generative models in geometry and albedo modeling, we aim to recover materials from high-quality geometry and base-colored 3D meshes.

However, extracting and recovering material representations for 3D meshes is challenging. Unlike previous material recognition methods \cite{lopes2023material, boss2020two, drehwald2023one}, this difficulty is heightened when identifying and separating different material regions within 3D objects in constricted albedo textures. These maps only reflect the base color and can be distorted, as shown by the globe in the upper right corner of \Cref{fig:teaser}. Additionally, shadows and lighting can affect judgment. Thus, the model must have strong material recognition capabilities and prior knowledge of object types and materials.

The emergence of Multimodal Large Language Models (MLLMs) \cite{Brown2020LanguageMA,openai2023gpt,Chowdhery2022PaLMSL,Anil2023PaLM2T,Hoffmann2022TrainingCL,Touvron2023LLaMAOA} provides novel approaches to problem-solving. These models have powerful visual understanding and recognition capabilities, along with a vast repository of prior knowledge, which covers the task of material estimation. Specifically, we are using GPT-4V(ision) for matching of materials. We first create highly detailed descriptions for materials to build a comprehensive library. Next, we use GPT-4V to retrieve materials for each segmented part of the object, utilizing visual prompts\cite{yang2023setofmark} and hierarchical text prompts. Finally, we meticulously designed algorithms to generate SVBRDF maps with consistent albedo, achieving realistic visual quality.

Notably, our work differs from the aforementioned studies by leveraging prior knowledge from foundation models like GPT-4V to extract and infer materials in albedo-only constrained scenarios. Additionally, we utilize existing material libraries as references to generate corresponding SVBRDF maps on a designed region-to-pixel algorithm. As illustrated in \Cref{fig:teaser}, our approach features: \textbf{1)} Enhanced 3D mesh realism: Leveraging GPT-4V's visual perception and external knowledge, our method improves the realism, depth, and visual quality of a wide range of mature 3D content generation models. \textbf{2)} Part-specific material matching: Ensuring material consistency with a segmentation network and refinement process, enabling precise material property retrieval for each segment. \textbf{3)} Rendering engine compatibility: Generating six comprehensive material maps(roughness, metallic, specular, normal, displacement, height), which are compatible with downstream rendering engines. Developers only need to paint albedo textures; material properties are then automatically generated, saving extensive time on detailed ambient occlusion masks and material map creation.

In summary, our contributions are as follows:

\begin{itemize}
  \item We present the first exploration of leveraging multimodal large language models (MLLMs), \eg GPT-4V for material recognition and unleashing their potential in applying real-world materials to extensive 3D objects with albedo-only.
  \item We create a material library containing thousands of materials with highly detailed descriptions readily for MLLMs to look up and assign.
  \item We develope an effective pipeline for texture segmentation, material matching and SVBRDF maps generation, enabling the high-quality application of materials to 3D assets.
\end{itemize}

\section{Related Work}
\noindent \textbf{3D Object Generation.}
The generation of 3D models using deep learning methods has experienced rapid development in recent times. The mainstream research can be primarily divided into two categories. The first category relies on techniques that optimize a Neural Radiance Field (NeRF)~\cite{mildenhall2021nerf} or 3D Gaussian~\cite{kerbl3Dgaussians} guided by 2D diffusion model through score-distillation-sampling(SDS) loss~\cite{poole2022dreamfusion, metzer2022latent, Lin_2023_CVPR, chen2023fantasia3d, wang2023prolificdreamer, shi2023mvdream, Tang2023DreamGaussianGG,wu2023hyperdreamer}. The second category aims at obtaining 3D representation via direct inference model, \eg, Point-E~\cite{nichol2022point}, Shap-E~\cite{jun2023shap}, and LRM~\cite{hong2023lrm}, proven fast with high quality through large-scale pretraining~\cite{hong20243dtopia,openlrm,tang2024lgm,li2023instant3d,xu2023dmv3d,tochilkin2024triposr,wang2023pflrm,wei2024meshlrm}.  
Although the capabilities of these methods are continuously improving, they still lack a high degree of realism in textures. More importantly, since the textures of 3D objects obtained by these methods are shaded, they cannot directly respond to lighting changes under different lighting conditions, leading to less realism. Although some works attempt to generate PBR textures, their results are considerably limited to generate physically realistic materials due to the low robustness of SDS~\cite{youwang2023paint,chen2023fantasia3d}. Our work is the first to introduce the prior knowledge of MLLMs into the texture synthesis process. We verifies that our method can be seamlessly and effectively applied to generated 3D objects, facilitating their downstream applications under different lighting condition.

\noindent \textbf{Material Capture and Generation.}
Recent studies such as Material Palette\cite{lopes2023material}, MatSim\cite{drehwald2023one}, TwoshotBRDF\cite{boss2020two} have made progress in the recognition and extraction of 3D object materials, allowing for the retrieval of SVBRDF information from images of real materials\cite{lopes2023material} and combining object shape and illumination \cite{boss2020two}, but they fail to extract and infer the materials of 3D objects with only albedo. On the other hand, works like Paint-it\cite{youwang2023paint}, Matlaber\cite{xu2023matlaber}, Collaborative\cite{vainer2024collaborative}, Fantasia3D\cite{chen2023fantasia3d}, and TANGO\cite{chen2022tango} focus on generating text-controllable 3D meshes with physically-based rendering material properties. However, these methods require either extensive training time on BRDF datasets or long inference time for generating materials. Additionally, they are limited to synthesizing only a subset of PBR textures and cannot generate the full range of material properties, such as height and displacement, which are essential for the fine tactile perception of object surfaces.

\noindent \textbf{Multimodal Large Language Models.} In the wake of the advancements achieved by large language models (LLMs)~\cite{Brown2020LanguageMA,openai2023gpt,Chowdhery2022PaLMSL,Anil2023PaLM2T,Hoffmann2022TrainingCL,Touvron2023LLaMAOA}, domain of research has increasingly turned its attention towards multimodal large language models (MLLMs). Recent advances in this field focus on the integration of vision understanding capabilities with LLMs~\cite{zhang2023internlmxcomposer,Alayrac2022FlamingoAV,Li2023BLIP2BL,Li2022BLIPBL,Huang2023LanguageIN,Driess2023PaLMEAE,Awadalla2023OpenFlamingoAO,dong2024internlmxcomposer2,sun2023alphaclip,qi2023gpt4point,chen2023sharegpt4v}.The advent of GPT-4V~\cite{2023GPT4VisionSC} has marked a significant milestone in the evolution of MLLMs, demonstrating groundbreaking 2D comprehending capabilities and open-world knowledge. Although GPT-4V cannot directly process 3D data, a pioneering work , GPTEval3D~\cite{wu2023gpteval3d}, first exploited GPT-4V's ability in evaluating the quality of generated 3D objects, and found that GPT-4V's judgement was in line with human evaluation. In this work, we delve into a new application of GPT-4V for material assignment of 3D objects.

\begin{figure*}[t]
  \centering
  \includegraphics[width=0.99\linewidth]{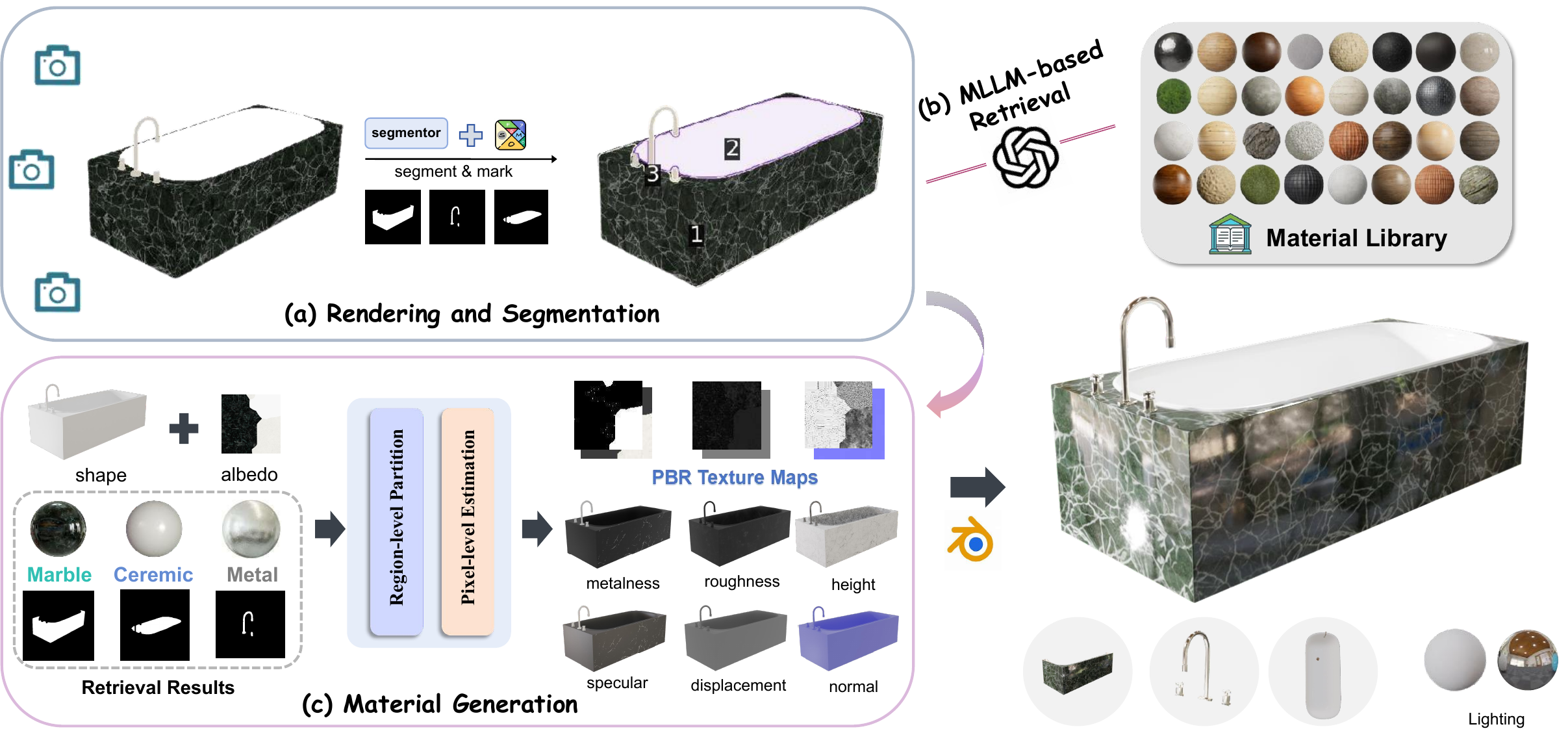}
   \vspace{-2pt}
   \caption{\textbf{Overall pipeline.} This pipeline of Make-it-Real is composed of image rendering and material segmentation, MLLM-based material retrieval, and SVBRDF Maps Generation. We finally use blender engine to conduct physically-based rendering.}
   \vspace{-15pt}
   \label{fig:overview}
\end{figure*}

\section{Methodology} 

\subsection{Preliminary}

Physically Based Rendering (PBR) materials are a compact representation of the bi-directional reflectance distribution function (BRDF), which describes how light is reflected from a surface. PBR material maps primarily encompass seven attributes: \textbf{A}lbedo, \textbf{R}oughness, \textbf{M}etallic, \textbf{N}ormal, \textbf{S}pecular, \textbf{H}eight, and \textbf{D}isplacement. Based on the rendering equation \cite{kajiya1986rendering}, given a location \( x \) and the surface normal \( n \), the incident light intensity at this point is denoted as \( L_i (\omega_i; x) \) along the direction \( \omega_i \); BRDF \( f_r (\omega_o, \omega_i; x) \) denotes the reflectance coefficient of the material viewing from direction \( \omega_o \). The observed light intensity \( L_o (\omega_o; x) \) is calculated over the hemisphere \( \Omega = \{ \omega_i : \omega_i \cdot n > 0 \} \):

\vspace{-5pt}

\begin{equation}
\label{eq:rendering_equation}
L_o (\omega_o; x) = \int_{\Omega} L_i (\omega_i; x) f_r (\omega_o, \omega_i; x) (\omega_i \cdot n) d\omega_i.
\end{equation}

Given the advancements in generating high-quality 3D shapes with albedo maps, the restoration of realistic material properties remains a challenge. We highlight a novel problem: given a 3D mesh (\(\tilde{O}\)) and known Albedo (\(\tilde{A}\)) map, which reflect the object's intrinsic appearance, the goal is to extract and restore all other SVBRDF attributes of the object, i.e. \textit{Make-it-Real}(\(\tilde{O}, \tilde{A}\)) = \(\{R, M, N, S, H, D\}\). Our setup supports the popular Cook-Torrance analytical BRDF model \cite{cook1982reflectance}. In this parameterization, the BRDF includes components for albedo \( b_a \in \mathbb{R}^3 \), metallic \( b_m \in \mathbb{R} \), and roughness \( b_r \in \mathbb{R} \). For more complex surface simulations, such as displacement and height modeling, we use the Blender rendering engine to simulate the BRDF function \( f_r (\omega_o, \omega_i; x) \).

\subsection{Make-it-Real: A Framework for Material Matching and Generataion}

In this section, we outline our material matching and generation pipeline, illustrated in \Cref{fig:overview}, which encompasses three stages: rendering and segmenting 3D meshes, retrieving matching materials using MLLM, and generating spatially varying BRDF maps from coarse to fine.

\subsubsection{Rendering and Material Segmentation}
\label{3_2_1:render_and_seg}

To accurately segment different material regions on 3D meshes with albedo maps, we propose an innovative segmentation strategy based on 2D image rendering in \Cref{fig:overview} (a). Initially, we use rasterization to render the input albedo mesh from various viewpoints to obtain a series of images:

\begin{equation}
\label{eq:rasterize}
\mathcal{I}(x, y) = \mathcal{R}\left(\text{UV}_{\text{map}}\left(\text{rasterize}(\tilde{O}, v_t, x, y)\right), \mathcal{T}\right)
\end{equation}

where \(\mathcal{I}(x, y)\) is the pixel value at image coordinates \((x, y)\), \(\text{rasterize}(\tilde{O}, v_t, \cdot)\) maps the 3D mesh \(\tilde{O}\) from viewpoint $v_t$ to 2D screen coordinates, \(\text{UV}_{\text{map}}(\cdot)\) converts rasterized coordinates to UV coordinates, \(\mathcal{T}\) is the albedo map, and \(\mathcal{R}(\cdot, \mathcal{T})\) samples color from the \(\mathcal{T}\) using the UV coordinates.

For the rendered images, we employ the Semantic-SAM \cite{li2023semanticsam} to perform preliminary semantic segmentation. Empirically, we select the main viewpoint with the largest projected area of the mesh, as it is more likely to contain more details. To address potential over-segmentation, drawing inspiration from \cite{ying2023omniseg3d}, we extract non-overlapping segments from the masks to form distinct patches, as detailed in our approach in \Cref{fig:mask_refine} (a). These patches are then merged based on similar colors to obtain the final material grouping. We incorporate Set-of-Mark \cite{yang2023setofmark} method to annotate each material segment with a unique identifier, sorted by area size from largest to smallest. This annotation acts as a visual cue, enhancing the visual comprehension capabilities of large multimodal language models.

\begin{figure*}[t]
  \centering
  \includegraphics[width=0.94\linewidth]{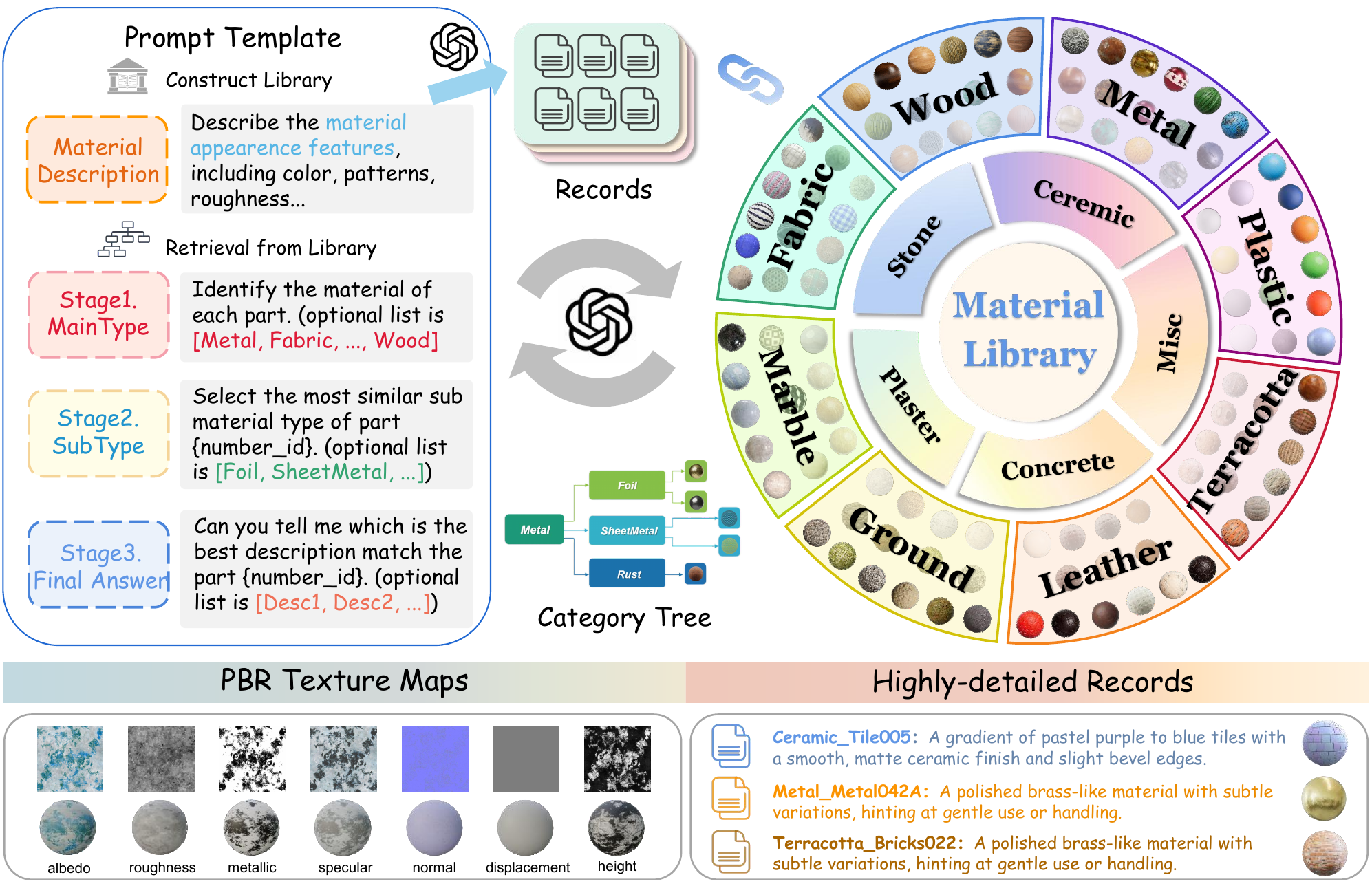}
   \vspace{-1pt}
   \caption{\textbf{The process of MLLM retrieving materials from the Material Library.} Utilizing GPT-4V model, we develop a material library, meticulously generating and cataloging comprehensive descriptions for each material. This structured repository facilitates hierarchical querying for material allocation in subsequent looking up processes.}
   \vspace{-15pt}
   \label{fig:mllm_ret}
\end{figure*}

\subsubsection{MLLM-based Material Retrieval}

\paragraph{Material library with fine-grained annotations.}
To enable large multimodal language models to accurately retrieve and match materials, we construct a finely annotated material library, as shown in \Cref{fig:mllm_ret}. It is composed of three main components: comprehensive PBR texture maps, highly-detailed records, and a category tree. It comprises 1,400 unique, tileable materials spanning 13 primary categories and 80 subtypes. The data primarily derives from the \cite{vecchio2024matsynth}, which offers comprehensive PBR material textures under a CC0 license with 4K resolution. Each material is represented by seven maps: albedo, roughness, metallic, specular, normal, displacement and height. Accompanying each material are highly-detailed annotations by GPT-4V, offering thorough descriptions of the material's visual characteristics and rich semantic information for the subsequent retrieval process. Created by crawling material sphere images and constructing prompts, these annotations capture subtle differences between materials, facilitating precise retrieval by GPT-4V, as detailed in \Cref{sup_b_3_prompts_detail}.

\paragraph{Hierarchical prompting for material retrieval.} 
Due to the vast size of our material library, feeding all prompts to GPT-4V simultaneously proves inefficient and challenging for memory retention. To ensure efficient and accurate material allocation in segmented areas of 3D meshes, we adopt a hierarchical text prompting approach. The schematic of the designed prompt is shown on the left side of \Cref{fig:mllm_ret}, and a complete querying process unfolds in \Cref{sup_b_3_prompts_detail}. This method starts by identifying the primary material types corresponding to each labeled region. Subsequently, hierarchical prompts guide GPT-4V to distinguish specific subclasses within the main material categories. We retrieve all descriptions under these subclasses to ascertain the most fitting descriptions for the segmented blocks. This hierarchical processing enables a more granular search of our material library, identifying the optimal description for each material segment. This approach aids in assigning the most suitable materials to each region and reduces memory and time consumption.

\subsubsection{SVBRDF Maps Generation}
\label{sec:3.2.3}
We propose a method to generate SVBRDF maps on a region-to-pixel scale in \Cref{fig:overview} (c). Initially, we segment texture map in uv space based on queried material regions on 2D image space. We then estimate BRDF values in pixel space using the object's original albedo map for reference, ensuring consistency with the albedo map. This approach effectively enhances the realism of rendered surfaces.

\begin{figure*}[t]
  \centering
  \includegraphics[width=0.9\linewidth]{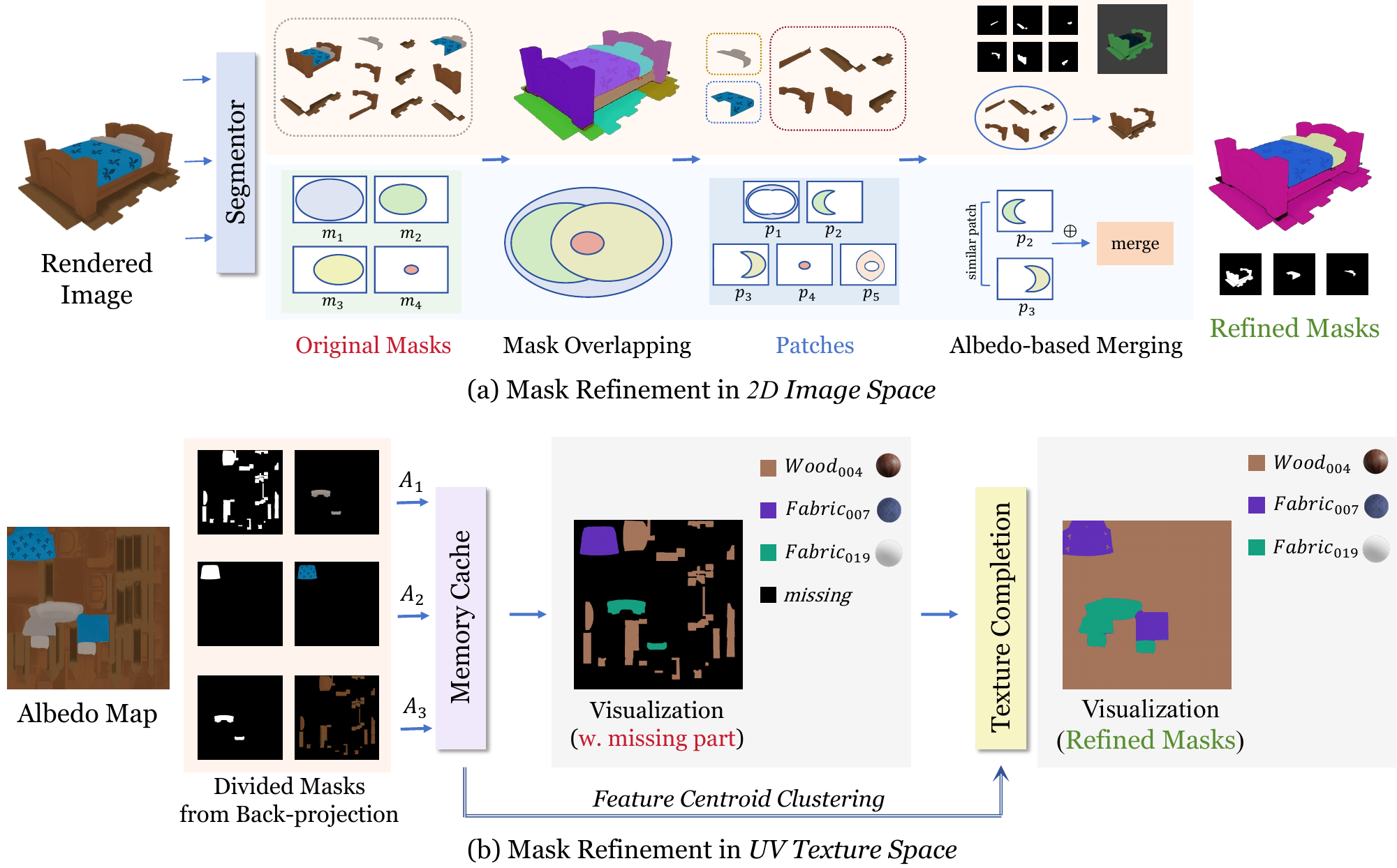}
  \vspace{-5pt}
   \caption{\textbf{Illustrations of mask refinement in 2D image space and UV texture space.} \textbf{(a)} We effectively cluster concise material-aware masks compared to original segmented parts from \cite{li2023semanticsam}. \textbf{(b)} We fix missing parts on the uv texture space to get a complete texture partition map.}
   \vspace{-15pt}
   \label{fig:mask_refine}
\end{figure*}

\paragraph{Region-level texture map partitioning.} 
Upon acquiring segmentation masks for material regions within the 2D rendered space of a 3D mesh, our objective transitions to transposing these segmentations from the 2D image space to the corresponding UV space. As described in  \Cref{3_2_1:render_and_seg},we extract 2D image features from 3D mesh points via rasterization, and then we apply the material masks to these features, facilitating the accurate transfer of segmentation to UV space. To project the image feature $I_t$ back to the texture atlas $\mathcal{T}_t$ with segmented image mask $m_t$ at the viewpoint $v_t$, we apply gradient-based optimization for $L_t$ over the values of $\mathcal{T}_t$ when rendered through the differential renderer $\mathcal{R}$, as presented in \Cref{eq:texture_projection}. In \Cref{eq:diff_cal}, we then compute the difference between $\mathcal{T}^{mask}_{t}$ and the initialized $\mathcal{T}_t$ to transfer the mask image feature $m_t$ into the texture space, represented by $m_{uv}$. The term $\sigma$ denotes the difference coefficient. Due to the limited perspectives available in rendering, we avoid using a naive median-filling approach to ensure that there are no missing areas on the texture map. Instead, we employ a block-centric clustering based on albedo, as illustrated in \Cref{fig:mask_refine} (b), to obtain cohesive and refined region masks. The process is shown in \Cref{sup_b_1_texture_partition}

\vspace{-15pt}
\begin{gather}
\nabla_{\mathcal{T}_t} L_t = \left( \mathcal{R}(\text{mesh}, \mathcal{T}_t, v_t) - I_t \right) \odot m_t \frac{\partial \mathcal{R}}{\partial \mathcal{T}_t}. \label{eq:texture_projection} \\
m_{uv} = \mathbf{1}\left(\sum_{i=1}^3 |\mathcal{T}^{mask}_{t} - \mathcal{T}_t| > \sigma \right). \label{eq:diff_cal}
\end{gather}

\paragraph{Pixel-level albedo-referenced estimation.} To achieve precise estimation of spatially varying BRDF (SVBRDF) at the pixel level, we draw inspiration from techniques commonly utilized by artists in creating texture maps. Artists typically use albedo maps as a reference for constructing ambient occlusion masks and further generating SVBRDF maps for material properties, which occupies a significant time portion of appearance modeling. Our method involves using the albedo map of the original object as a reference to refine the retrieved materials. 
We enhance the querying process using a KD-Tree algorithm, which searches for the nearest neighbor pixel index in the key albedo(retrieved map) for each RGB value of the queried albedo(input map) pixel, detailed in \Cref{sup_b_2_albedo_refer}. This process ensures that areas with similar colors exhibit similar BRDF values, avoiding abrupt changes in material properties; For regions with greater color differences, the distribution of differences aligns consistently with the input albedo, to simulate variations such as embossments or scratches. We  retrieves SVBRDF values at pixel level, maintaining texture consistency with the albedo and producing appropriate concave or smooth surface effects. We further analyze the effects in \Cref{sec:refine_obja}.

\begin{figure*}[t]
  \centering
  \includegraphics[width=1.0\linewidth]{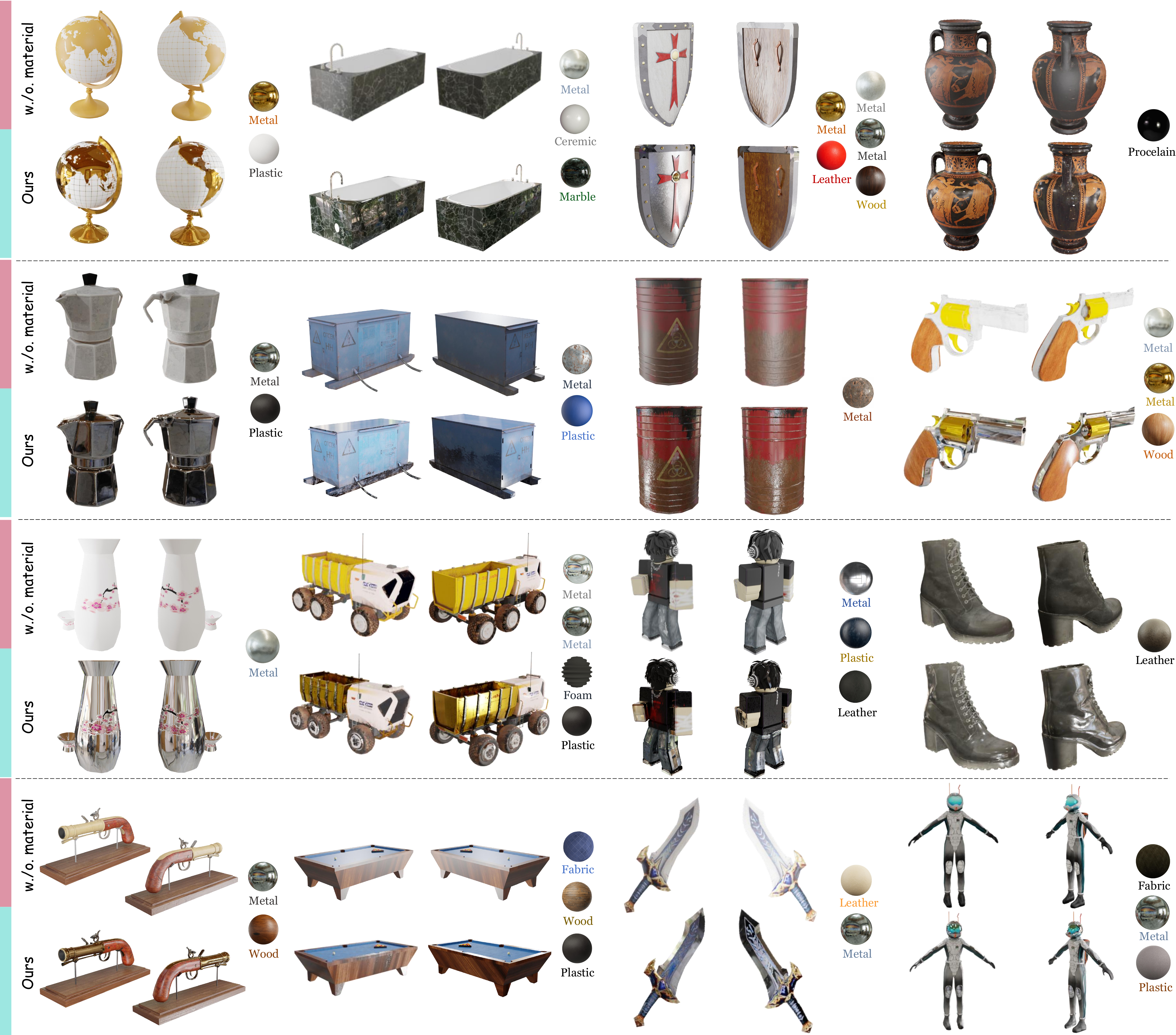}
  \vspace{-15pt}
   \caption{\textbf{Qualitative results of \methodname refining 3D asserts without PBR maps}. Objects are selected from Objaverse~\cite{deitke2022objaverse} with albedo maps only.}
   \vspace{-15pt}
   \label{fig:objaverse}
\end{figure*}

\vspace{5pt}

\begin{figure*}[!t]
  \centering
  \includegraphics[width=0.92\linewidth]{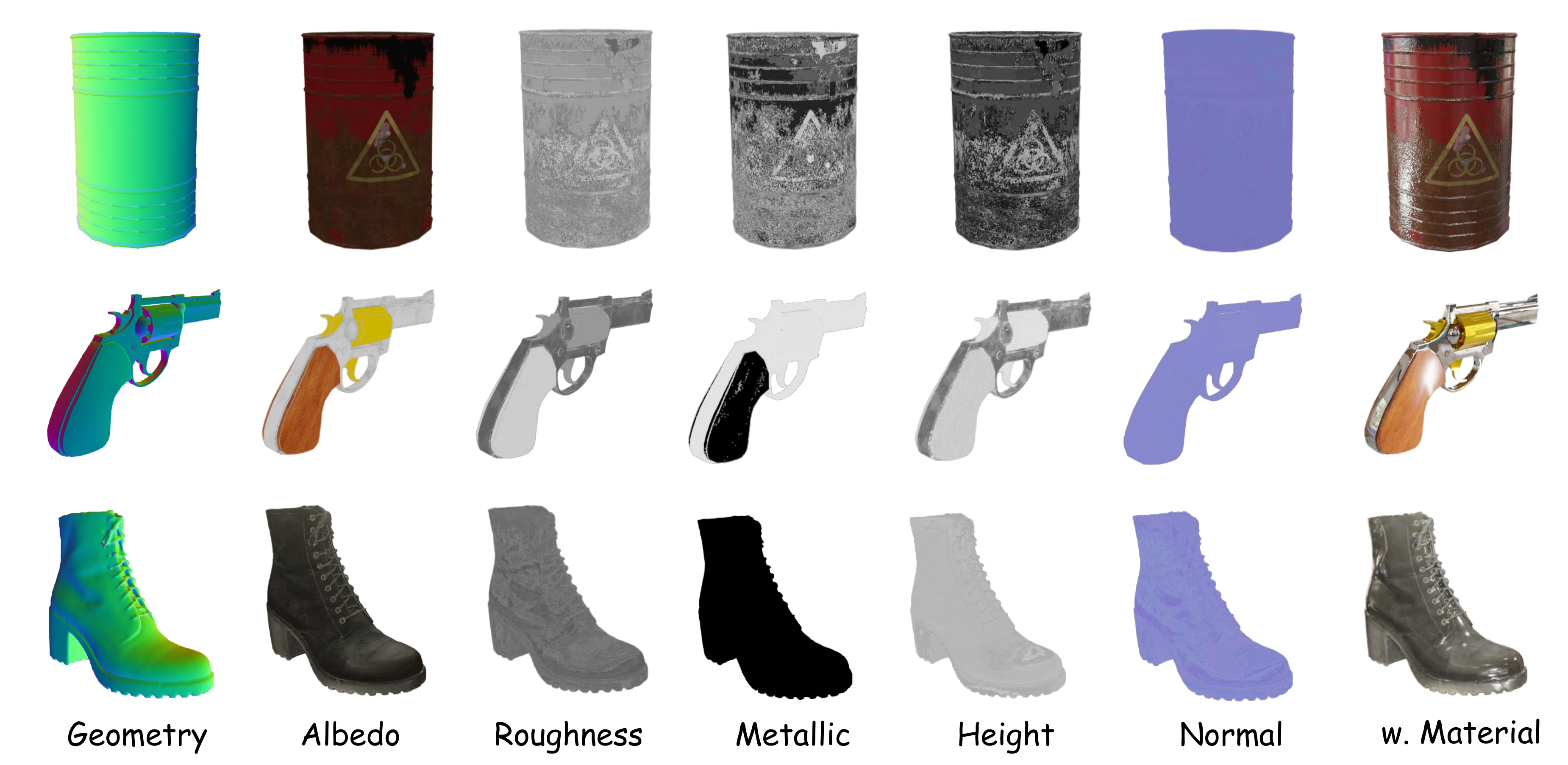}
  \vspace{-1pt}
   \caption{\textbf{Visualization of generated texture maps}. We visualize some SVBRDF maps, where the material maps are well aligned with the albedo maps.}
   \vspace{-20pt}
   \label{fig:tex_map}
\end{figure*}

\begin{figure*}[!t]
  \centering
  \includegraphics[width=\linewidth]{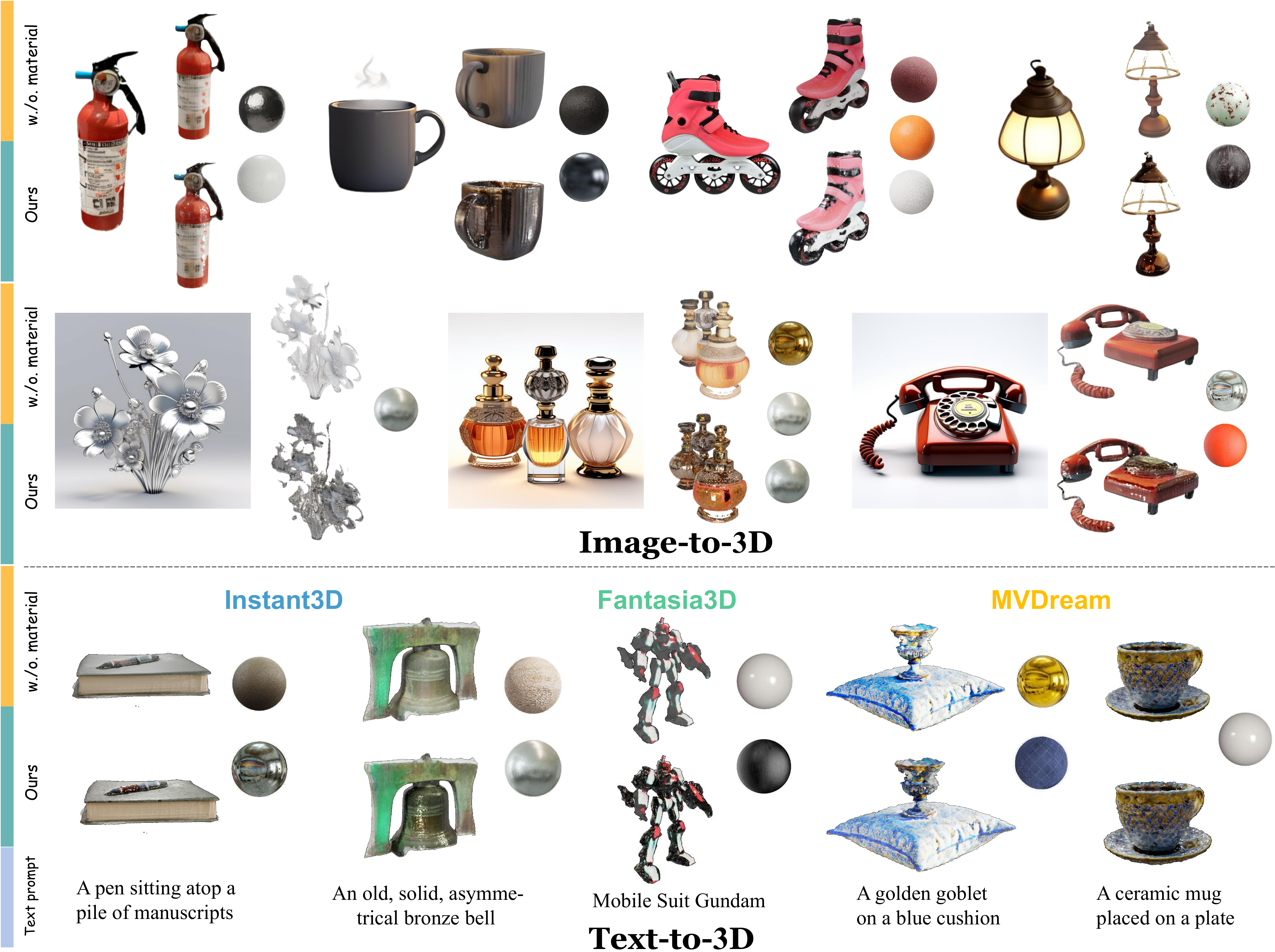}
   \vspace{-1pt}
   \caption{\textbf{Qualitative comparisons} between \methodname refining results and 3D objects generated by edge-cutting 3D content creation models. The upper row depicts image-to-3D models (InstantMesh and TripoSR), and the lower row shows results of text-to-3D models.} 
   \vspace{-15pt}
   \label{fig:generated_3d_refine}
\end{figure*}

\section{Experiments}
\label{sec:exp}

\subsection{Experiment settings}

To verify the effectiveness of \methodname, we conduct refinement experiments mainly on two types of objects. The first type is artificial 3D assets, with the primary model from Objaverse~\cite{deitke2022objaverse}; The second type is objects generated by state-of-the-art 3D generation methods. For existing 3D assets, we pick 200 objects with diverse textures from Objaverse by human experts. For 3D generative models (InstantMesh~\cite{xu2024instantmesh}, TripoSR~\cite{tochilkin2024triposr}, MVDream~\cite{shi2023mvdream}, Instant3D~\cite{instant3d2023} and Fantasia3D~\cite{chen2023fantasia3d}), we also generate 200 objects for each methods use prompts designed in GPTEval3D~\cite{wu2023gpteval3d}. We use \methodname to refine the objects and compare the texture quality before and after refinement. We perform both GPT-4V~\cite{2023GPT4VisionSC} based evaluation and user study on the above objects (Detailed guidence and prompts are available in ~\Cref{sup:gpt_evaluation}).

\subsection{Experiment results}

\textbf{Texture refinement for existing 3D assets.}
\label{sec:refine_obja}
As shown in \Cref{fig:objaverse}, assets processed through \methodname demonstrate the capability to accurately segment objects, assign various suitable materials, and synthesize high-fidelity, photorealistic textures. Some materials exhibit notable highlights, such as a marble bathtub and a globe made of gold. The interaction of these different materials with light varies significantly, leading to diverse reflective effects under the same environmental lighting, thus presenting a range of textures. Additionally, material properties vary across different regions; for instance, the globe's landmass and handle exhibit gold characteristics, while other parts are identified as plastic. Furthermore, the texture and appearance of materials also vary, such as the subtle wrinkles on the red box in the second column and the more pronounced color contrast at the base of the blue box, which enhances realism and reflects signs of use. Due to the albedo-referenced algorithm design in \Cref{sec:3.2.3}, regions with similar colors have similar BRDF values, avoiding abrupt changes in material properties, such as the continuous gold surface of the globe and the silver body of the kettle. Regions with significant color differences display consistent differential distributions with the key map, such as the embossed textures on the lower left corner of the water bottle in \Cref{fig:objaverse} and the subtle particle variations on the surface of the red oil drum. Additionally, visualization of texture maps demonstrates reasonably consistent textures with albedo, shown in \Cref{fig:tex_map}. Quantitive results are shown in \Cref{tab:gpt_eval}, and more qualitative results can be found in \Cref{sup:more_results}.

\textbf{Texture refinement for generated 3D objects.} \Cref{fig:generated_3d_refine} displays the qualitative results.
By leveraging our model for enhancement, we observe that \methodname successfully generates appropriate material maps for both image-to-3D and text-to-3D models. 

Similarly, our \methodname successfully paint these models with materials. As reported in \Cref{tab:gpt_eval}, human experts consistently favor the objects post-refinement across all evaluated 3D content creation methods. This preference aligns with the evaluations performed by GPT-4V, indicating a general consensus on the enhancement in quality achieved through \methodname refinement process.

\begin{table}[t]
  \centering
  \caption{\textbf{GPT evaluation and user preference}. GPT's and user's preference comparison on \methodname refined objects sourced from existing 3D assets and state-of-the-art 3D generation methods.}
  \small
    \begin{tabular}{lccccc}
    \toprule
    \multirow{2}[0]{*}{Domain} & \multirow{2}[0]{*}{Method / Source} & \multicolumn{2}{c}{GPT Evaluation} & \multicolumn{2}{c}{User Preference} \\
          &       & Base object & +Make-it-Real & Base object & +Make-it-Real \\
    \midrule
    3D assets & Objaverse~\cite{deitke2022objaverse} & 15.2$\%$ & 84.8$\%$ & 22.2$\%$ & 77.8$\%$ \\
    \midrule
\multirow{2}[0]{*}{Image-to-3D} & InstantMesh~\cite{xu2024instantmesh} & 28.2$\%$  & 71.8$\%$  &  31.1$\%$  & 68.9$\%$ \\
          & TripoSR~\cite{tochilkin2024triposr} & 36.4$\%$  & 63.6$\%$  &   33.0$\%$   & 77.0$\%$  \\
    \midrule
    \multirow{3}[0]{*}{Text-to-3D} & Instant3D~\cite{li2023instant3d} & 38.5$\%$  & 61.5$\%$  &  35.4$\%$  & 64.6$\%$  \\
          & MVDream~\cite{shi2023mvdream} & 44.1$\%$  & 55.9$\%$  &  41.5$\%$  & 58.5$\%$ \\
          & Fantasia3D~\cite{chen2023fantasia3d} & 46.2$\%$  & 53.8$\%$  & 48.7$\%$   &  51.3$\%$ \\
    \bottomrule
    \end{tabular}%
    
  \label{tab:gpt_eval}%
  \vspace{-8pt}
\end{table}%

\begin{figure*}[t]
  \centering
  \begin{minipage}{0.92\linewidth}
    \centering
    \includegraphics[width=\linewidth]{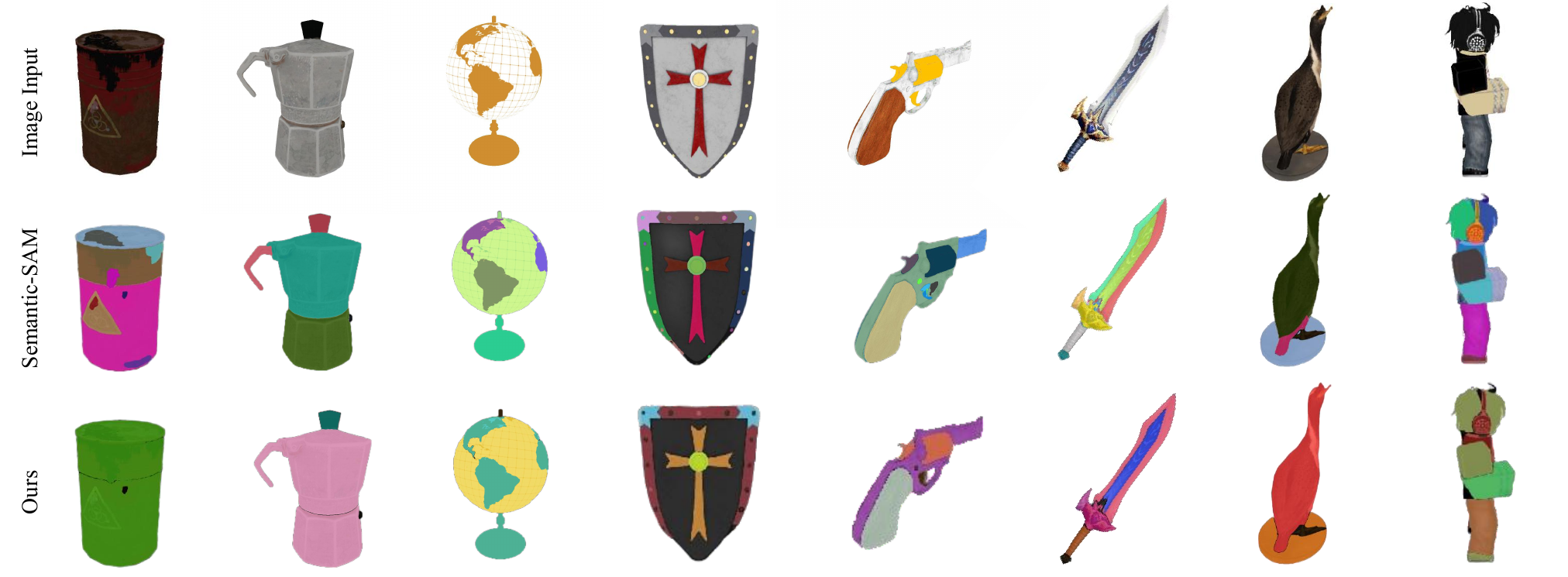}
    \vspace{-12pt}
    \caption{\textbf{Ablation study of material segmentation refinement.} Compared to direct usage of SemanticSAM~\cite{li2023semanticsam}, Our post-process tailored for material segmentation on 3D object can produce more consistent results.}
    \label{fig:ablation_refinement_1}
  \end{minipage}
  \vspace{14pt} 
  \begin{minipage}{0.92\linewidth}
    \centering
    \includegraphics[width=\linewidth]{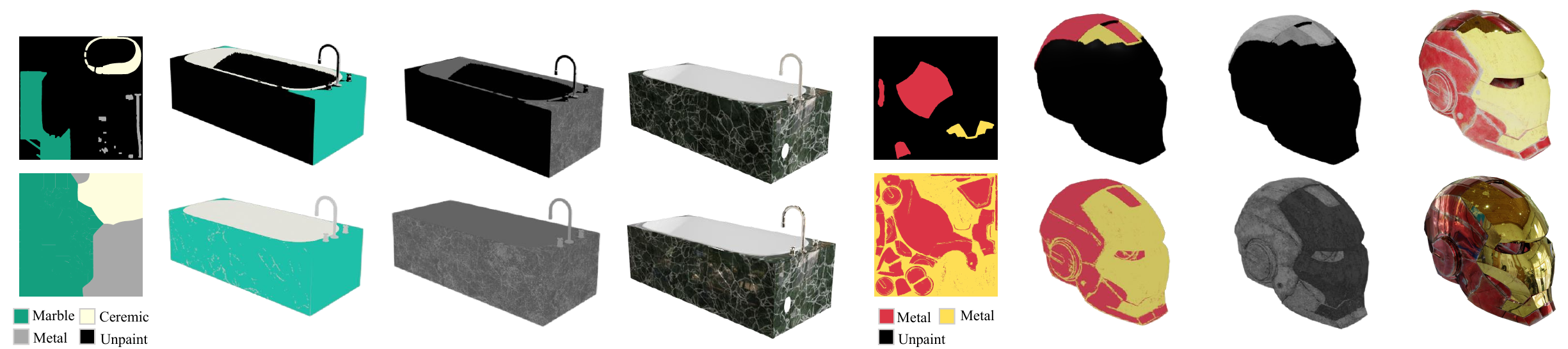}
    \vspace{-12pt}
    \caption{\textbf{Ablation study of missing part refinement.} Our method on the bottom row produces consistent texture maps and avoids the missing parts of material texture.}
    \label{fig:ablation_refinement_2}
  \end{minipage}
  \vspace{-14pt}
\end{figure*}
\vspace{-10pt}

\subsection{Ablation Study}

\textbf{Effects of Mask Refinement Module.} In \Cref{3_2_1:render_and_seg}, we performed additional material post-processing on the segmentation outcomes, as depicted in \Cref{fig:ablation_refinement_1}. The refined results indicate that our module achieves precise material segmentation for most standard objects, thereby enabling more accurate queries by GPT-4V. In \Cref{sec:3.2.3}, we addressed the completion of missing regions in the texture map within the UV space at the regional level, as illustrated in \Cref{fig:ablation_refinement_2}. This method not only increases texture coverage but also enhances the visual quality and consistency of the refined 3D model. More ablation studies can be found in \Cref{sup_d_1}.

\section{Conclusion}
In this paper, we present a novel framework leveraging MLLMs prior of the world to build a material library and proposing an automatic pipeline to refine and synthesize new PBR maps for initial 3D models, achieving highly photo-realistic PBR textures maps synthesis. Experimental results confirm that our approach can automatically refine both generated and CAD models to achieve photo-realism under dynamic lighting conditions. We believe \methodname is a new and promising solution in the last few procedures of AI based 3D content creation pipeline with the development of MLLMs like GPT-4V~\cite{2023GPT4VisionSC} as well as the roaring field of deep learning based 3D generation from scratch.

%
%
\bibliographystyle{plainnat}
\bibliography{main}

\begin{thebibliography}{69}
\providecommand{\natexlab}[1]{#1}
\providecommand{\url}[1]{\texttt{#1}}
\expandafter\ifx\csname urlstyle\endcsname\relax
  \providecommand{\doi}[1]{doi: #1}\else
  \providecommand{\doi}{doi: \begingroup \urlstyle{rm}\Url}\fi

\bibitem[Alayrac et~al.(2022)Alayrac, Donahue, Luc, Miech, Barr, Hasson, Lenc, Mensch, Millican, Reynolds, Ring, Rutherford, Cabi, Han, Gong, Samangooei, Monteiro, Menick, Borgeaud, Brock, Nematzadeh, Sharifzadeh, Binkowski, Barreira, Vinyals, Zisserman, and Simonyan]{Alayrac2022FlamingoAV}
Jean-Baptiste Alayrac, Jeff Donahue, Pauline Luc, Antoine Miech, Iain Barr, Yana Hasson, Karel Lenc, Arthur Mensch, Katie Millican, Malcolm Reynolds, Roman Ring, Eliza Rutherford, Serkan Cabi, Tengda Han, Zhitao Gong, Sina Samangooei, Marianne Monteiro, Jacob Menick, Sebastian Borgeaud, Andy Brock, Aida Nematzadeh, Sahand Sharifzadeh, Mikolaj Binkowski, Ricardo Barreira, Oriol Vinyals, Andrew Zisserman, and Karen Simonyan.
\newblock Flamingo: a visual language model for few-shot learning.
\newblock \emph{ArXiv}, abs/2204.14198, 2022.
\newblock URL \url{https://api.semanticscholar.org/CorpusID:248476411}.

\bibitem[Anil et~al.(2023)Anil, Dai, Firat, Johnson, Lepikhin, Passos, Shakeri, Taropa, Bailey, Chen, Chu, Clark, Shafey, Huang, Meier-Hellstern, Mishra, Moreira, Omernick, Robinson, Ruder, Tay, Xiao, Xu, Zhang, Abrego, Ahn, Austin, Barham, Botha, Bradbury, Brahma, Brooks, Catasta, Cheng, Cherry, Choquette-Choo, Chowdhery, Cr{\'e}py, Dave, Dehghani, Dev, Devlin, D'iaz, Du, Dyer, Feinberg, Feng, Fienber, Freitag, Garc{\'i}a, Gehrmann, Gonz{\'a}lez, Gur-Ari, Hand, Hashemi, Hou, Howland, Hu, Hui, Hurwitz, Isard, Ittycheriah, Jagielski, Jia, Kenealy, Krikun, Kudugunta, Lan, Lee, Lee, Li, Li, Li, Li, Li, Lim, Lin, Liu, Liu, Maggioni, Mahendru, Maynez, Misra, Moussalem, Nado, Nham, Ni, Nystrom, Parrish, Pellat, Polacek, Polozov, Pope, Qiao, Reif, Richter, Riley, Ros, Roy, Saeta, Samuel, Shelby, Slone, Smilkov, So, Sohn, Tokumine, Valter, Vasudevan, Vodrahalli, Wang, Wang, Wang, Wang, Wieting, Wu, Xu, Xu, Xue, Yin, Yu, Zhang, Zheng, Zheng, Zhou, Zhou, Petrov, and Wu]{Anil2023PaLM2T}
Rohan Anil, Andrew~M. Dai, Orhan Firat, Melvin Johnson, Dmitry Lepikhin, Alexandre~Tachard Passos, Siamak Shakeri, Emanuel Taropa, Paige Bailey, Z.~Chen, Eric Chu, J.~Clark, Laurent~El Shafey, Yanping Huang, Kathleen~S. Meier-Hellstern, Gaurav Mishra, Erica Moreira, Mark Omernick, Kevin Robinson, Sebastian Ruder, Yi~Tay, Kefan Xiao, Yuanzhong Xu, Yujing Zhang, Gustavo~Hernandez Abrego, Junwhan Ahn, Jacob Austin, Paul Barham, Jan~A. Botha, James Bradbury, Siddhartha Brahma, Kevin~Michael Brooks, Michele Catasta, Yongzhou Cheng, Colin Cherry, Christopher~A. Choquette-Choo, Aakanksha Chowdhery, C~Cr{\'e}py, Shachi Dave, Mostafa Dehghani, Sunipa Dev, Jacob Devlin, M.~C. D'iaz, Nan Du, Ethan Dyer, Vladimir Feinberg, Fan Feng, Vlad Fienber, Markus Freitag, Xavier Garc{\'i}a, Sebastian Gehrmann, Lucas Gonz{\'a}lez, Guy Gur-Ari, Steven Hand, Hadi Hashemi, Le~Hou, Joshua Howland, An~Ren Hu, Jeffrey Hui, Jeremy Hurwitz, Michael Isard, Abe Ittycheriah, Matthew Jagielski, Wen~Hao Jia, Kathleen Kenealy, Maxim Krikun,
  Sneha Kudugunta, Chang Lan, Katherine Lee, Benjamin Lee, Eric Li, Mu-Li Li, Wei Li, Yaguang Li, Jun~Yu Li, Hyeontaek Lim, Han Lin, Zhong-Zhong Liu, Frederick Liu, Marcello Maggioni, Aroma Mahendru, Joshua Maynez, Vedant Misra, Maysam Moussalem, Zachary Nado, John Nham, Eric Ni, Andrew Nystrom, Alicia Parrish, Marie Pellat, Martin Polacek, Alex Polozov, Reiner Pope, Siyuan Qiao, Emily Reif, Bryan Richter, Parker Riley, Alexandra Ros, Aurko Roy, Brennan Saeta, Rajkumar Samuel, Renee~Marie Shelby, Ambrose Slone, Daniel Smilkov, David~R. So, Daniela Sohn, Simon Tokumine, Dasha Valter, Vijay Vasudevan, Kiran Vodrahalli, Xuezhi Wang, Pidong Wang, Zirui Wang, Tao Wang, John Wieting, Yuhuai Wu, Ke~Xu, Yunhan Xu, Lin~Wu Xue, Pengcheng Yin, Jiahui Yu, Qiaoling Zhang, Steven Zheng, Ce~Zheng, Wei Zhou, Denny Zhou, Slav Petrov, and Yonghui Wu.
\newblock Palm 2 technical report.
\newblock \emph{ArXiv}, abs/2305.10403, 2023.
\newblock URL \url{https://api.semanticscholar.org/CorpusID:258740735}.

\bibitem[Awadalla et~al.(2023)Awadalla, Gao, Gardner, Hessel, Hanafy, Zhu, Marathe, Bitton, Gadre, Sagawa, Jitsev, Kornblith, Koh, Ilharco, Wortsman, and Schmidt]{Awadalla2023OpenFlamingoAO}
Anas Awadalla, Irena Gao, Josh Gardner, Jack Hessel, Yusuf Hanafy, Wanrong Zhu, Kalyani Marathe, Yonatan Bitton, Samir~Yitzhak Gadre, Shiori Sagawa, Jenia Jitsev, Simon Kornblith, Pang~Wei Koh, Gabriel Ilharco, Mitchell Wortsman, and Ludwig Schmidt.
\newblock Openflamingo: An open-source framework for training large autoregressive vision-language models.
\newblock \emph{ArXiv}, abs/2308.01390, 2023.
\newblock URL \url{https://api.semanticscholar.org/CorpusID:261043320}.

\bibitem[Boss et~al.(2020)Boss, Jampani, Kim, Lensch, and Kautz]{boss2020two}
Mark Boss, Varun Jampani, Kihwan Kim, Hendrik Lensch, and Jan Kautz.
\newblock Two-shot spatially-varying brdf and shape estimation.
\newblock In \emph{Proceedings of the IEEE/CVF Conference on Computer Vision and Pattern Recognition}, pages 3982--3991, 2020.

\bibitem[Brown et~al.(2020)Brown, Mann, Ryder, Subbiah, Kaplan, Dhariwal, Neelakantan, Shyam, Sastry, Askell, Agarwal, Herbert-Voss, Krueger, Henighan, Child, Ramesh, Ziegler, Wu, Winter, Hesse, Chen, Sigler, Litwin, Gray, Chess, Clark, Berner, McCandlish, Radford, Sutskever, and Amodei]{Brown2020LanguageMA}
Tom~B. Brown, Benjamin Mann, Nick Ryder, Melanie Subbiah, Jared Kaplan, Prafulla Dhariwal, Arvind Neelakantan, Pranav Shyam, Girish Sastry, Amanda Askell, Sandhini Agarwal, Ariel Herbert-Voss, Gretchen Krueger, T.~J. Henighan, Rewon Child, Aditya Ramesh, Daniel~M. Ziegler, Jeff Wu, Clemens Winter, Christopher Hesse, Mark Chen, Eric Sigler, Mateusz Litwin, Scott Gray, Benjamin Chess, Jack Clark, Christopher Berner, Sam McCandlish, Alec Radford, Ilya Sutskever, and Dario Amodei.
\newblock Language models are few-shot learners.
\newblock \emph{ArXiv}, abs/2005.14165, 2020.
\newblock URL \url{https://api.semanticscholar.org/CorpusID:218971783}.

\bibitem[Chan et~al.(2021)Chan, Lin, Chan, Nagano, Pan, Mello, Gallo, Guibas, Tremblay, Khamis, Karras, and Wetzstein]{Chan2021EfficientG3}
Eric Chan, Connor~Z. Lin, Matthew Chan, Koki Nagano, Boxiao Pan, Shalini~De Mello, Orazio Gallo, Leonidas~J. Guibas, Jonathan Tremblay, S.~Khamis, Tero Karras, and Gordon Wetzstein.
\newblock Efficient geometry-aware 3d generative adversarial networks.
\newblock \emph{2022 IEEE/CVF Conference on Computer Vision and Pattern Recognition (CVPR)}, pages 16102--16112, 2021.
\newblock URL \url{https://api.semanticscholar.org/CorpusID:245144673}.

\bibitem[Chen et~al.(2023{\natexlab{a}})Chen, Li, Dong, Zhang, He, Wang, Zhao, and Lin]{chen2023sharegpt4v}
Lin Chen, Jisong Li, Xiaoyi Dong, Pan Zhang, Conghui He, Jiaqi Wang, Feng Zhao, and Dahua Lin.
\newblock Sharegpt4v: Improving large multi-modal models with better captions.
\newblock \emph{arXiv preprint arXiv:2311.12793}, 2023{\natexlab{a}}.

\bibitem[Chen et~al.(2023{\natexlab{b}})Chen, Chen, Jiao, and Jia]{Chen2023Fantasia3DDG}
Rui Chen, Y.~Chen, Ningxin Jiao, and Kui Jia.
\newblock Fantasia3d: Disentangling geometry and appearance for high-quality text-to-3d content creation.
\newblock \emph{2023 IEEE/CVF International Conference on Computer Vision (ICCV)}, pages 22189--22199, 2023{\natexlab{b}}.
\newblock URL \url{https://api.semanticscholar.org/CorpusID:257757213}.

\bibitem[Chen et~al.(2023{\natexlab{c}})Chen, Chen, Jiao, and Jia]{chen2023fantasia3d}
Rui Chen, Yongwei Chen, Ningxin Jiao, and Kui Jia.
\newblock Fantasia3d: Disentangling geometry and appearance for high-quality text-to-3d content creation.
\newblock \emph{arXiv preprint arXiv:2303.13873}, 2023{\natexlab{c}}.

\bibitem[Chen et~al.(2022{\natexlab{a}})Chen, Jiang, Song, Yang, Black, Geiger, and Hilliges]{chen2022gdna}
Xu~Chen, Tianjian Jiang, Jie Song, Jinlong Yang, Michael~J Black, Andreas Geiger, and Otmar Hilliges.
\newblock gdna: Towards generative detailed neural avatars.
\newblock In \emph{Proceedings of the IEEE/CVF Conference on Computer Vision and Pattern Recognition}, pages 20427--20437, 2022{\natexlab{a}}.

\bibitem[Chen et~al.(2022{\natexlab{b}})Chen, Chen, Lei, Zhang, and Jia]{chen2022tango}
Yongwei Chen, Rui Chen, Jiabao Lei, Yabin Zhang, and Kui Jia.
\newblock Tango: Text-driven photorealistic and robust 3d stylization via lighting decomposition.
\newblock \emph{Advances in Neural Information Processing Systems}, 35:\penalty0 30923--30936, 2022{\natexlab{b}}.

\bibitem[Chowdhery et~al.(2022)Chowdhery, Narang, Devlin, Bosma, Mishra, Roberts, Barham, Chung, Sutton, Gehrmann, Schuh, Shi, Tsvyashchenko, Maynez, Rao, Barnes, Tay, Shazeer, Prabhakaran, Reif, Du, Hutchinson, Pope, Bradbury, Austin, Isard, Gur-Ari, Yin, Duke, Levskaya, Ghemawat, Dev, Michalewski, Garc{\'i}a, Misra, Robinson, Fedus, Zhou, Ippolito, Luan, Lim, Zoph, Spiridonov, Sepassi, Dohan, Agrawal, Omernick, Dai, Pillai, Pellat, Lewkowycz, Moreira, Child, Polozov, Lee, Zhou, Wang, Saeta, D{\'i}az, Firat, Catasta, Wei, Meier-Hellstern, Eck, Dean, Petrov, and Fiedel]{Chowdhery2022PaLMSL}
Aakanksha Chowdhery, Sharan Narang, Jacob Devlin, Maarten Bosma, Gaurav Mishra, Adam Roberts, Paul Barham, Hyung~Won Chung, Charles Sutton, Sebastian Gehrmann, Parker Schuh, Kensen Shi, Sasha Tsvyashchenko, Joshua Maynez, Abhishek Rao, Parker Barnes, Yi~Tay, Noam~M. Shazeer, Vinodkumar Prabhakaran, Emily Reif, Nan Du, Benton~C. Hutchinson, Reiner Pope, James Bradbury, Jacob Austin, Michael Isard, Guy Gur-Ari, Pengcheng Yin, Toju Duke, Anselm Levskaya, Sanjay Ghemawat, Sunipa Dev, Henryk Michalewski, Xavier Garc{\'i}a, Vedant Misra, Kevin Robinson, Liam Fedus, Denny Zhou, Daphne Ippolito, David Luan, Hyeontaek Lim, Barret Zoph, Alexander Spiridonov, Ryan Sepassi, David Dohan, Shivani Agrawal, Mark Omernick, Andrew~M. Dai, Thanumalayan~Sankaranarayana Pillai, Marie Pellat, Aitor Lewkowycz, Erica Moreira, Rewon Child, Oleksandr Polozov, Katherine Lee, Zongwei Zhou, Xuezhi Wang, Brennan Saeta, Mark D{\'i}az, Orhan Firat, Michele Catasta, Jason Wei, Kathleen~S. Meier-Hellstern, Douglas Eck, Jeff Dean, Slav Petrov,
  and Noah Fiedel.
\newblock Palm: Scaling language modeling with pathways.
\newblock \emph{J. Mach. Learn. Res.}, 24:\penalty0 240:1--240:113, 2022.
\newblock URL \url{https://api.semanticscholar.org/CorpusID:247951931}.

\bibitem[Cook and Torrance(1982)]{cook1982reflectance}
Robert~L Cook and Kenneth~E. Torrance.
\newblock A reflectance model for computer graphics.
\newblock \emph{ACM Transactions on Graphics (ToG)}, 1\penalty0 (1):\penalty0 7--24, 1982.

\bibitem[Deitke et~al.(2022{\natexlab{a}})Deitke, Schwenk, Salvador, Weihs, Michel, VanderBilt, Schmidt, Ehsani, Kembhavi, and Farhadi]{deitke2022objaverse}
Matt Deitke, Dustin Schwenk, Jordi Salvador, Luca Weihs, Oscar Michel, Eli VanderBilt, Ludwig Schmidt, Kiana Ehsani, Aniruddha Kembhavi, and Ali Farhadi.
\newblock Objaverse: A universe of annotated 3d objects, 2022{\natexlab{a}}.

\bibitem[Deitke et~al.(2022{\natexlab{b}})Deitke, Schwenk, Salvador, Weihs, Michel, VanderBilt, Schmidt, Ehsani, Kembhavi, and Farhadi]{objaverse}
Matt Deitke, Dustin Schwenk, Jordi Salvador, Luca Weihs, Oscar Michel, Eli VanderBilt, Ludwig Schmidt, Kiana Ehsani, Aniruddha Kembhavi, and Ali Farhadi.
\newblock Objaverse: A universe of annotated 3d objects.
\newblock \emph{arXiv preprint arXiv:2212.08051}, 2022{\natexlab{b}}.

\bibitem[Deitke et~al.(2023)Deitke, Liu, Wallingford, Ngo, Michel, Kusupati, Fan, Laforte, Voleti, Gadre, VanderBilt, Kembhavi, Vondrick, Gkioxari, Ehsani, Schmidt, and Farhadi]{objaverseXL}
Matt Deitke, Ruoshi Liu, Matthew Wallingford, Huong Ngo, Oscar Michel, Aditya Kusupati, Alan Fan, Christian Laforte, Vikram Voleti, Samir~Yitzhak Gadre, Eli VanderBilt, Aniruddha Kembhavi, Carl Vondrick, Georgia Gkioxari, Kiana Ehsani, Ludwig Schmidt, and Ali Farhadi.
\newblock Objaverse-xl: A universe of 10m+ 3d objects.
\newblock \emph{arXiv preprint arXiv:2307.05663}, 2023.

\bibitem[Dong et~al.(2024)Dong, Zhang, Zang, Cao, Wang, Ouyang, Wei, Zhang, Duan, Cao, Zhang, Li, Yan, Gao, Zhang, Li, Li, Chen, He, Zhang, Qiao, Lin, and Wang]{dong2024internlmxcomposer2}
Xiaoyi Dong, Pan Zhang, Yuhang Zang, Yuhang Cao, Bin Wang, Linke Ouyang, Xilin Wei, Songyang Zhang, Haodong Duan, Maosong Cao, Wenwei Zhang, Yining Li, Hang Yan, Yang Gao, Xinyue Zhang, Wei Li, Jingwen Li, Kai Chen, Conghui He, Xingcheng Zhang, Yu~Qiao, Dahua Lin, and Jiaqi Wang.
\newblock Internlm-xcomposer2: Mastering free-form text-image composition and comprehension in vision-language large model, 2024.

\bibitem[Dong et~al.(2023)Dong, Chen, Yang, Black, Hilliges, and Geiger]{dong2023ag3d}
Zijian Dong, Xu~Chen, Jinlong Yang, Michael~J Black, Otmar Hilliges, and Andreas Geiger.
\newblock Ag3d: Learning to generate 3d avatars from 2d image collections.
\newblock \emph{arXiv preprint arXiv:2305.02312}, 2023.

\bibitem[Drehwald et~al.(2023)Drehwald, Eppel, Li, Hao, and Aspuru-Guzik]{drehwald2023one}
Manuel~S Drehwald, Sagi Eppel, Jolina Li, Han Hao, and Alan Aspuru-Guzik.
\newblock One-shot recognition of any material anywhere using contrastive learning with physics-based rendering.
\newblock In \emph{Proceedings of the IEEE/CVF International Conference on Computer Vision}, pages 23524--23533, 2023.

\bibitem[Driess et~al.(2023)Driess, Xia, Sajjadi, Lynch, Chowdhery, Ichter, Wahid, Tompson, Vuong, Yu, Huang, Chebotar, Sermanet, Duckworth, Levine, Vanhoucke, Hausman, Toussaint, Greff, Zeng, Mordatch, and Florence]{Driess2023PaLMEAE}
Danny Driess, F.~Xia, Mehdi S.~M. Sajjadi, Corey Lynch, Aakanksha Chowdhery, Brian Ichter, Ayzaan Wahid, Jonathan Tompson, Quan~Ho Vuong, Tianhe Yu, Wenlong Huang, Yevgen Chebotar, Pierre Sermanet, Daniel Duckworth, Sergey Levine, Vincent Vanhoucke, Karol Hausman, Marc Toussaint, Klaus Greff, Andy Zeng, Igor Mordatch, and Peter~R. Florence.
\newblock Palm-e: An embodied multimodal language model.
\newblock In \emph{International Conference on Machine Learning}, 2023.
\newblock URL \url{https://api.semanticscholar.org/CorpusID:257364842}.

\bibitem[He and Wang(2023)]{openlrm}
Zexin He and Tengfei Wang.
\newblock Openlrm: Open-source large reconstruction models.
\newblock \url{https://github.com/3DTopia/OpenLRM}, 2023.

\bibitem[Hoffmann et~al.(2022)Hoffmann, Borgeaud, Mensch, Buchatskaya, Cai, Rutherford, de~Las~Casas, Hendricks, Welbl, Clark, Hennigan, Noland, Millican, van~den Driessche, Damoc, Guy, Osindero, Simonyan, Elsen, Rae, Vinyals, and Sifre]{Hoffmann2022TrainingCL}
Jordan Hoffmann, Sebastian Borgeaud, Arthur Mensch, Elena Buchatskaya, Trevor Cai, Eliza Rutherford, Diego de~Las~Casas, Lisa~Anne Hendricks, Johannes Welbl, Aidan Clark, Tom Hennigan, Eric Noland, Katie Millican, George van~den Driessche, Bogdan Damoc, Aurelia Guy, Simon Osindero, Karen Simonyan, Erich Elsen, Jack~W. Rae, Oriol Vinyals, and L.~Sifre.
\newblock Training compute-optimal large language models.
\newblock \emph{ArXiv}, abs/2203.15556, 2022.
\newblock URL \url{https://api.semanticscholar.org/CorpusID:247778764}.

\bibitem[Hong et~al.(2024)Hong, Tang, Cao, Shi, Wu, Chen, Wang, Pan, Lin, and Liu]{hong20243dtopia}
Fangzhou Hong, Jiaxiang Tang, Ziang Cao, Min Shi, Tong Wu, Zhaoxi Chen, Tengfei Wang, Liang Pan, Dahua Lin, and Ziwei Liu.
\newblock 3dtopia: Large text-to-3d generation model with hybrid diffusion priors, 2024.

\bibitem[Hong et~al.(2023)Hong, Zhang, Gu, Bi, Zhou, Liu, Liu, Sunkavalli, Bui, and Tan]{hong2023lrm}
Yicong Hong, Kai Zhang, Jiuxiang Gu, Sai Bi, Yang Zhou, Difan Liu, Feng Liu, Kalyan Sunkavalli, Trung Bui, and Hao Tan.
\newblock Lrm: Large reconstruction model for single image to 3d, 2023.

\bibitem[Huang et~al.(2023)Huang, Dong, Wang, Hao, Singhal, Ma, Lv, Cui, Mohammed, Liu, Aggarwal, Chi, Bjorck, Chaudhary, Som, Song, and Wei]{Huang2023LanguageIN}
Shaohan Huang, Li~Dong, Wenhui Wang, Yaru Hao, Saksham Singhal, Shuming Ma, Tengchao Lv, Lei Cui, Owais~Khan Mohammed, Qiang Liu, Kriti Aggarwal, Zewen Chi, Johan Bjorck, Vishrav Chaudhary, Subhojit Som, Xia Song, and Furu Wei.
\newblock Language is not all you need: Aligning perception with language models.
\newblock \emph{ArXiv}, abs/2302.14045, 2023.
\newblock URL \url{https://api.semanticscholar.org/CorpusID:257219775}.

\bibitem[Jatavallabhula et~al.(2019)Jatavallabhula, Smith, Lafleche, Tsang, Rozantsev, Chen, Xiang, Lebaredian, and Fidler]{jatavallabhula2019kaolin}
Krishna~Murthy Jatavallabhula, Edward Smith, Jean-Francois Lafleche, Clement~Fuji Tsang, Artem Rozantsev, Wenzheng Chen, Tommy Xiang, Rev Lebaredian, and Sanja Fidler.
\newblock Kaolin: A pytorch library for accelerating 3d deep learning research, 2019.

\bibitem[Jun and Nichol(2023)]{jun2023shap}
Heewoo Jun and Alex Nichol.
\newblock Shap-e: Generating conditional 3d implicit functions.
\newblock \emph{arXiv preprint arXiv:2305.02463}, 2023.

\bibitem[Kajiya(1986)]{kajiya1986rendering}
James~T Kajiya.
\newblock The rendering equation.
\newblock In \emph{Proceedings of the 13th annual conference on Computer graphics and interactive techniques}, pages 143--150, 1986.

\bibitem[Kerbl et~al.(2023)Kerbl, Kopanas, Leimk{\"u}hler, and Drettakis]{kerbl3Dgaussians}
Bernhard Kerbl, Georgios Kopanas, Thomas Leimk{\"u}hler, and George Drettakis.
\newblock 3d gaussian splatting for real-time radiance field rendering.
\newblock \emph{ACM Transactions on Graphics}, 42\penalty0 (4), July 2023.
\newblock URL \url{https://repo-sam.inria.fr/fungraph/3d-gaussian-splatting/}.

\bibitem[Labsch{\"u}tz et~al.(2011)Labsch{\"u}tz, Kr{\"o}sl, Aquino, Grash{\"a}ftl, and Kohl]{labschutz2011content}
Matthias Labsch{\"u}tz, Katharina Kr{\"o}sl, Mariebeth Aquino, Florian Grash{\"a}ftl, and Stephanie Kohl.
\newblock Content creation for a 3d game with maya and unity 3d.
\newblock \emph{Institute of Computer Graphics and Algorithms, Vienna University of Technology}, 6:\penalty0 124, 2011.

\bibitem[Li et~al.(2023{\natexlab{a}})Li, Zhang, Sun, Zou, Liu, Yang, Li, Zhang, and Gao]{li2023semanticsam}
Feng Li, Hao Zhang, Peize Sun, Xueyan Zou, Shilong Liu, Jianwei Yang, Chunyuan Li, Lei Zhang, and Jianfeng Gao.
\newblock Semantic-sam: Segment and recognize anything at any granularity, 2023{\natexlab{a}}.

\bibitem[Li et~al.(2023{\natexlab{b}})Li, Tan, Zhang, Xu, Luan, Xu, Hong, Sunkavalli, Shakhnarovich, and Bi]{instant3d2023}
Jiahao Li, Hao Tan, Kai Zhang, Zexiang Xu, Fujun Luan, Yinghao Xu, Yicong Hong, Kalyan Sunkavalli, Greg Shakhnarovich, and Sai Bi.
\newblock Instant3d: Fast text-to-3d with sparse-view generation and large reconstruction model.
\newblock \emph{https://arxiv.org/abs/2311.06214}, 2023{\natexlab{b}}.

\bibitem[Li et~al.(2023{\natexlab{c}})Li, Tan, Zhang, Xu, Luan, Xu, Hong, Sunkavalli, Shakhnarovich, and Bi]{li2023instant3d}
Jiahao Li, Hao Tan, Kai Zhang, Zexiang Xu, Fujun Luan, Yinghao Xu, Yicong Hong, Kalyan Sunkavalli, Greg Shakhnarovich, and Sai Bi.
\newblock Instant3d: Fast text-to-3d with sparse-view generation and large reconstruction model, 2023{\natexlab{c}}.

\bibitem[Li et~al.(2022)Li, Li, Xiong, and Hoi]{Li2022BLIPBL}
Junnan Li, Dongxu Li, Caiming Xiong, and Steven C.~H. Hoi.
\newblock Blip: Bootstrapping language-image pre-training for unified vision-language understanding and generation.
\newblock In \emph{International Conference on Machine Learning}, 2022.
\newblock URL \url{https://api.semanticscholar.org/CorpusID:246411402}.

\bibitem[Li et~al.(2023{\natexlab{d}})Li, Li, Savarese, and Hoi]{Li2023BLIP2BL}
Junnan Li, Dongxu Li, Silvio Savarese, and Steven C.~H. Hoi.
\newblock Blip-2: Bootstrapping language-image pre-training with frozen image encoders and large language models.
\newblock \emph{ArXiv}, abs/2301.12597, 2023{\natexlab{d}}.
\newblock URL \url{https://api.semanticscholar.org/CorpusID:256390509}.

\bibitem[Li et~al.(2024)Li, Gan, Luo, Wang, Liu, Zhang, Li, Yin, Zhang, and Peng]{li2024materialseg3d}
Zeyu Li, Ruitong Gan, Chuanchen Luo, Yuxi Wang, Jiaheng Liu, Ziwei Zhu~Man Zhang, Qing Li, Xucheng Yin, Zhaoxiang Zhang, and Junran Peng.
\newblock Materialseg3d: Segmenting dense materials from 2d priors for 3d assets.
\newblock \emph{arXiv preprint arXiv:2404.13923}, 2024.

\bibitem[Lin et~al.(2023)Lin, Gao, Tang, Takikawa, Zeng, Huang, Kreis, Fidler, Liu, and Lin]{Lin_2023_CVPR}
Chen-Hsuan Lin, Jun Gao, Luming Tang, Towaki Takikawa, Xiaohui Zeng, Xun Huang, Karsten Kreis, Sanja Fidler, Ming-Yu Liu, and Tsung-Yi Lin.
\newblock Magic3d: High-resolution text-to-3d content creation.
\newblock In \emph{Proceedings of the IEEE/CVF Conference on Computer Vision and Pattern Recognition (CVPR)}, pages 300--309, June 2023.

\bibitem[Lopes et~al.(2023)Lopes, Pizzati, and de~Charette]{lopes2023material}
Ivan Lopes, Fabio Pizzati, and Raoul de~Charette.
\newblock Material palette: Extraction of materials from a single image.
\newblock \emph{arXiv preprint arXiv:2311.17060}, 2023.

\bibitem[Metzer et~al.(2022)Metzer, Richardson, Patashnik, Giryes, and Cohen-Or]{metzer2022latent}
Gal Metzer, Elad Richardson, Or~Patashnik, Raja Giryes, and Daniel Cohen-Or.
\newblock Latent-nerf for shape-guided generation of 3d shapes and textures.
\newblock \emph{arXiv preprint arXiv:2211.07600}, 2022.

\bibitem[Mildenhall et~al.(2021)Mildenhall, Srinivasan, Tancik, Barron, Ramamoorthi, and Ng]{mildenhall2021nerf}
Ben Mildenhall, Pratul~P Srinivasan, Matthew Tancik, Jonathan~T Barron, Ravi Ramamoorthi, and Ren Ng.
\newblock Nerf: Representing scenes as neural radiance fields for view synthesis.
\newblock \emph{Communications of the ACM}, 65\penalty0 (1):\penalty0 99--106, 2021.

\bibitem[Nichol et~al.(2022)Nichol, Jun, Dhariwal, Mishkin, and Chen]{nichol2022point}
Alex Nichol, Heewoo Jun, Prafulla Dhariwal, Pamela Mishkin, and Mark Chen.
\newblock Point-e: A system for generating 3d point clouds from complex prompts.
\newblock \emph{arXiv preprint arXiv:2212.08751}, 2022.

\bibitem[OpenAI(2023{\natexlab{a}})]{2023GPT4VisionSC}
OpenAI.
\newblock Gpt-4v(ision) system card.
\newblock \emph{OpenAI}, 2023{\natexlab{a}}.
\newblock URL \url{https://api.semanticscholar.org/CorpusID:263218031}.

\bibitem[OpenAI(2023{\natexlab{b}})]{openai2023gpt}
R~OpenAI.
\newblock Gpt-4 technical report.
\newblock \emph{arXiv}, pages 2303--08774, 2023{\natexlab{b}}.

\bibitem[Park et~al.(2018)Park, Rematas, Farhadi, and Seitz]{park2018photoshape}
Keunhong Park, Konstantinos Rematas, Ali Farhadi, and Steven~M Seitz.
\newblock Photoshape: Photorealistic materials for large-scale shape collections.
\newblock \emph{arXiv preprint arXiv:1809.09761}, 2018.

\bibitem[Poole et~al.(2022)Poole, Jain, Barron, and Mildenhall]{poole2022dreamfusion}
Ben Poole, Ajay Jain, Jonathan~T. Barron, and Ben Mildenhall.
\newblock Dreamfusion: Text-to-3d using 2d diffusion.
\newblock \emph{arXiv}, 2022.

\bibitem[Qi et~al.(2023)Qi, Fang, Sun, Wu, Wu, Wang, Lin, and Zhao]{qi2023gpt4point}
Zhangyang Qi, Ye~Fang, Zeyi Sun, Xiaoyang Wu, Tong Wu, Jiaqi Wang, Dahua Lin, and Hengshuang Zhao.
\newblock Gpt4point: A unified framework for point-language understanding and generation.
\newblock \emph{arXiv preprint arXiv:2312.02980}, 2023.

\bibitem[Richardson et~al.(2023)Richardson, Metzer, Alaluf, Giryes, and Cohen-Or]{richardson2023texture}
Elad Richardson, Gal Metzer, Yuval Alaluf, Raja Giryes, and Daniel Cohen-Or.
\newblock Texture: Text-guided texturing of 3d shapes.
\newblock \emph{arXiv preprint arXiv:2302.01721}, 2023.

\bibitem[Sharma et~al.(2023)Sharma, Philip, Gharbi, Freeman, Durand, and Deschaintre]{sharma2023materialistic}
Prafull Sharma, Julien Philip, Micha{\"e}l Gharbi, Bill Freeman, Fredo Durand, and Valentin Deschaintre.
\newblock Materialistic: Selecting similar materials in images.
\newblock \emph{ACM Transactions on Graphics (TOG)}, 42\penalty0 (4):\penalty0 1--14, 2023.

\bibitem[Shi et~al.(2023)Shi, Wang, Ye, Mai, Li, and Yang]{shi2023mvdream}
Yichun Shi, Peng Wang, Jianglong Ye, Long Mai, Kejie Li, and Xiao Yang.
\newblock Mvdream: Multi-view diffusion for 3d generation.
\newblock \emph{arXiv:2308.16512}, 2023.

\bibitem[Sun et~al.(2023)Sun, Fang, Wu, Zhang, Zang, Kong, Xiong, Lin, and Wang]{sun2023alphaclip}
Zeyi Sun, Ye~Fang, Tong Wu, Pan Zhang, Yuhang Zang, Shu Kong, Yuanjun Xiong, Dahua Lin, and Jiaqi Wang.
\newblock Alpha-clip: A clip model focusing on wherever you want, 2023.

\bibitem[Tang et~al.(2023)Tang, Ren, Zhou, Liu, and Zeng]{Tang2023DreamGaussianGG}
Jiaxiang Tang, Jiawei Ren, Hang Zhou, Ziwei Liu, and Gang Zeng.
\newblock Dreamgaussian: Generative gaussian splatting for efficient 3d content creation.
\newblock \emph{ArXiv}, abs/2309.16653, 2023.
\newblock URL \url{https://api.semanticscholar.org/CorpusID:263131552}.

\bibitem[Tang et~al.(2024)Tang, Chen, Chen, Wang, Zeng, and Liu]{tang2024lgm}
Jiaxiang Tang, Zhaoxi Chen, Xiaokang Chen, Tengfei Wang, Gang Zeng, and Ziwei Liu.
\newblock Lgm: Large multi-view gaussian model for high-resolution 3d content creation.
\newblock \emph{arXiv preprint arXiv:2402.05054}, 2024.

\bibitem[Tochilkin et~al.(2024)Tochilkin, Pankratz, Liu, Huang, Letts, Li, Liang, Laforte, Jampani, and Cao]{tochilkin2024triposr}
Dmitry Tochilkin, David Pankratz, Zexiang Liu, Zixuan Huang, Adam Letts, Yangguang Li, Ding Liang, Christian Laforte, Varun Jampani, and Yan-Pei Cao.
\newblock Triposr: Fast 3d object reconstruction from a single image, 2024.

\bibitem[Touvron et~al.(2023)Touvron, Lavril, Izacard, Martinet, Lachaux, Lacroix, Rozi{\`e}re, Goyal, Hambro, Azhar, Rodriguez, Joulin, Grave, and Lample]{Touvron2023LLaMAOA}
Hugo Touvron, Thibaut Lavril, Gautier Izacard, Xavier Martinet, Marie-Anne Lachaux, Timoth{\'e}e Lacroix, Baptiste Rozi{\`e}re, Naman Goyal, Eric Hambro, Faisal Azhar, Aurelien Rodriguez, Armand Joulin, Edouard Grave, and Guillaume Lample.
\newblock Llama: Open and efficient foundation language models.
\newblock \emph{ArXiv}, abs/2302.13971, 2023.
\newblock URL \url{https://api.semanticscholar.org/CorpusID:257219404}.

\bibitem[Vainer et~al.(2024)Vainer, Boss, Parger, Kutsy, De~Nigris, Rowles, Perony, and Donn{\'e}]{vainer2024collaborative}
Shimon Vainer, Mark Boss, Mathias Parger, Konstantin Kutsy, Dante De~Nigris, Ciara Rowles, Nicolas Perony, and Simon Donn{\'e}.
\newblock Collaborative control for geometry-conditioned pbr image generation.
\newblock \emph{arXiv preprint arXiv:2402.05919}, 2024.

\bibitem[Vecchio and Deschaintre(2024)]{vecchio2024matsynth}
Giuseppe Vecchio and Valentin Deschaintre.
\newblock Matsynth: A modern pbr materials dataset.
\newblock \emph{arXiv preprint arXiv:2401.06056}, 2024.

\bibitem[Wang and Shi(2023)]{wang2023imagedream}
Peng Wang and Yichun Shi.
\newblock Imagedream: Image-prompt multi-view diffusion for 3d generation, 2023.

\bibitem[Wang et~al.(2023{\natexlab{a}})Wang, Tan, Bi, Xu, Luan, Sunkavalli, Wang, Xu, and Zhang]{wang2023pflrm}
Peng Wang, Hao Tan, Sai Bi, Yinghao Xu, Fujun Luan, Kalyan Sunkavalli, Wenping Wang, Zexiang Xu, and Kai Zhang.
\newblock Pf-lrm: Pose-free large reconstruction model for joint pose and shape prediction, 2023{\natexlab{a}}.

\bibitem[Wang et~al.(2023{\natexlab{b}})Wang, Lu, Wang, Bao, Li, Su, and Zhu]{wang2023prolificdreamer}
Zhengyi Wang, Cheng Lu, Yikai Wang, Fan Bao, Chongxuan Li, Hang Su, and Jun Zhu.
\newblock Prolificdreamer: High-fidelity and diverse text-to-3d generation with variational score distillation.
\newblock \emph{arXiv preprint arXiv:2305.16213}, 2023{\natexlab{b}}.

\bibitem[Wei et~al.(2024)Wei, Zhang, Bi, Tan, Luan, Deschaintre, Sunkavalli, Su, and Xu]{wei2024meshlrm}
Xinyue Wei, Kai Zhang, Sai Bi, Hao Tan, Fujun Luan, Valentin Deschaintre, Kalyan Sunkavalli, Hao Su, and Zexiang Xu.
\newblock Meshlrm: Large reconstruction model for high-quality mesh, 2024.

\bibitem[Wu et~al.(2023)Wu, Li, Yang, Zhang, Pan, Wang, Lin, and Liu]{wu2023hyperdreamer}
Tong Wu, Zhibing Li, Shuai Yang, Pan Zhang, Xinggang Pan, Jiaqi Wang, Dahua Lin, and Ziwei Liu.
\newblock Hyperdreamer: Hyper-realistic 3d content generation and editing from a single image, 2023.

\bibitem[Wu et~al.(2024)Wu, Yang, Li, Zhang, Liu, Guibas, Lin, and Wetzstein]{wu2023gpteval3d}
Tong Wu, Guandao Yang, Zhibing Li, Kai Zhang, Ziwei Liu, Leonidas Guibas, Dahua Lin, and Gordon Wetzstein.
\newblock Gpt-4v (ision) is a human-aligned evaluator for text-to-3d generation.
\newblock \emph{arXiv preprint arXiv:2401.04092}, 2024.

\bibitem[Xu et~al.(2024)Xu, Cheng, Gao, Wang, Gao, and Shan]{xu2024instantmesh}
Jiale Xu, Weihao Cheng, Yiming Gao, Xintao Wang, Shenghua Gao, and Ying Shan.
\newblock Instantmesh: Efficient 3d mesh generation from a single image with sparse-view large reconstruction models.
\newblock \emph{arXiv preprint arXiv:2404.07191}, 2024.

\bibitem[Xu et~al.(2023{\natexlab{a}})Xu, Lyu, Pan, and Dai]{xu2023matlaber}
Xudong Xu, Zhaoyang Lyu, Xingang Pan, and Bo~Dai.
\newblock Matlaber: Material-aware text-to-3d via latent brdf auto-encoder.
\newblock \emph{arXiv preprint arXiv:2308.09278}, 2023{\natexlab{a}}.

\bibitem[Xu et~al.(2023{\natexlab{b}})Xu, Tan, Luan, Bi, Wang, Li, Shi, Sunkavalli, Wetzstein, Xu, and Zhang]{xu2023dmv3d}
Yinghao Xu, Hao Tan, Fujun Luan, Sai Bi, Peng Wang, Jiahao Li, Zifan Shi, Kalyan Sunkavalli, Gordon Wetzstein, Zexiang Xu, and Kai Zhang.
\newblock Dmv3d: Denoising multi-view diffusion using 3d large reconstruction model, 2023{\natexlab{b}}.

\bibitem[Yang et~al.(2023)Yang, Zhang, Li, Zou, Li, and Gao]{yang2023setofmark}
Jianwei Yang, Hao Zhang, Feng Li, Xueyan Zou, Chunyuan Li, and Jianfeng Gao.
\newblock Set-of-mark prompting unleashes extraordinary visual grounding in gpt-4v, 2023.

\bibitem[Ying et~al.(2023)Ying, Yin, Zhang, Wang, Yu, Huang, and Fang]{ying2023omniseg3d}
Haiyang Ying, Yixuan Yin, Jinzhi Zhang, Fan Wang, Tao Yu, Ruqi Huang, and Lu~Fang.
\newblock Omniseg3d: Omniversal 3d segmentation via hierarchical contrastive learning.
\newblock \emph{arXiv preprint arXiv:2311.11666}, 2023.

\bibitem[Youwang et~al.(2023)Youwang, Oh, and Pons-Moll]{youwang2023paint}
Kim Youwang, Tae-Hyun Oh, and Gerard Pons-Moll.
\newblock Paint-it: Text-to-texture synthesis via deep convolutional texture map optimization and physically-based rendering.
\newblock \emph{arXiv preprint arXiv:2312.11360}, 2023.

\bibitem[Zhang et~al.(2023)Zhang, Dong, Wang, Cao, Xu, Ouyang, Zhao, Duan, Zhang, Ding, Zhang, Yan, Zhang, Li, Li, Chen, He, Zhang, Qiao, Lin, and Wang]{zhang2023internlmxcomposer}
Pan Zhang, Xiaoyi Dong, Bin Wang, Yuhang Cao, Chao Xu, Linke Ouyang, Zhiyuan Zhao, Haodong Duan, Songyang Zhang, Shuangrui Ding, Wenwei Zhang, Hang Yan, Xinyue Zhang, Wei Li, Jingwen Li, Kai Chen, Conghui He, Xingcheng Zhang, Yu~Qiao, Dahua Lin, and Jiaqi Wang.
\newblock Internlm-xcomposer: A vision-language large model for advanced text-image comprehension and composition, 2023.

\end{thebibliography}

\clearpage
\appendix
\section{Supplementary Material Overview}
In this supplementary material, we provide additional details and results that are not included in the main paper due to the space limit. The attached video includes intuitive and interesting qualitative results of \methodname. 

\section{Details of Make-it-Real}
In this section, we detail the pipeline that briefly outlined in the main paper. We commence by elaborating on the generation of SVBRDF Maps, incorporating illustrative figures to detail the process involving operations in computer graphics. \Cref{sup_b_1_texture_partition} is dedicated to explaining the acquisition of region-level texture map partition. Following that, in \Cref{sup_b_2_albedo_refer} we discuss the method of pixel-level albedo-referenced estimation. Then, in \Cref{sup_b_m_setup} we declare some details of rendering procedure. \Cref{sup_b_3_prompts_detail} details the prompt design for material captioning and matching. \Cref{sup_b_4_evaluation_detail} reports the details of the GPT-4V based evaluation. Potential broader impacts are discussed in \Cref{sup:broader_impacts}.

\subsection{Texture Partition Module Design}
\label{sup_b_1_texture_partition}

\begin{figure*}[b]
  \centering
  \includegraphics[width=0.97\linewidth]{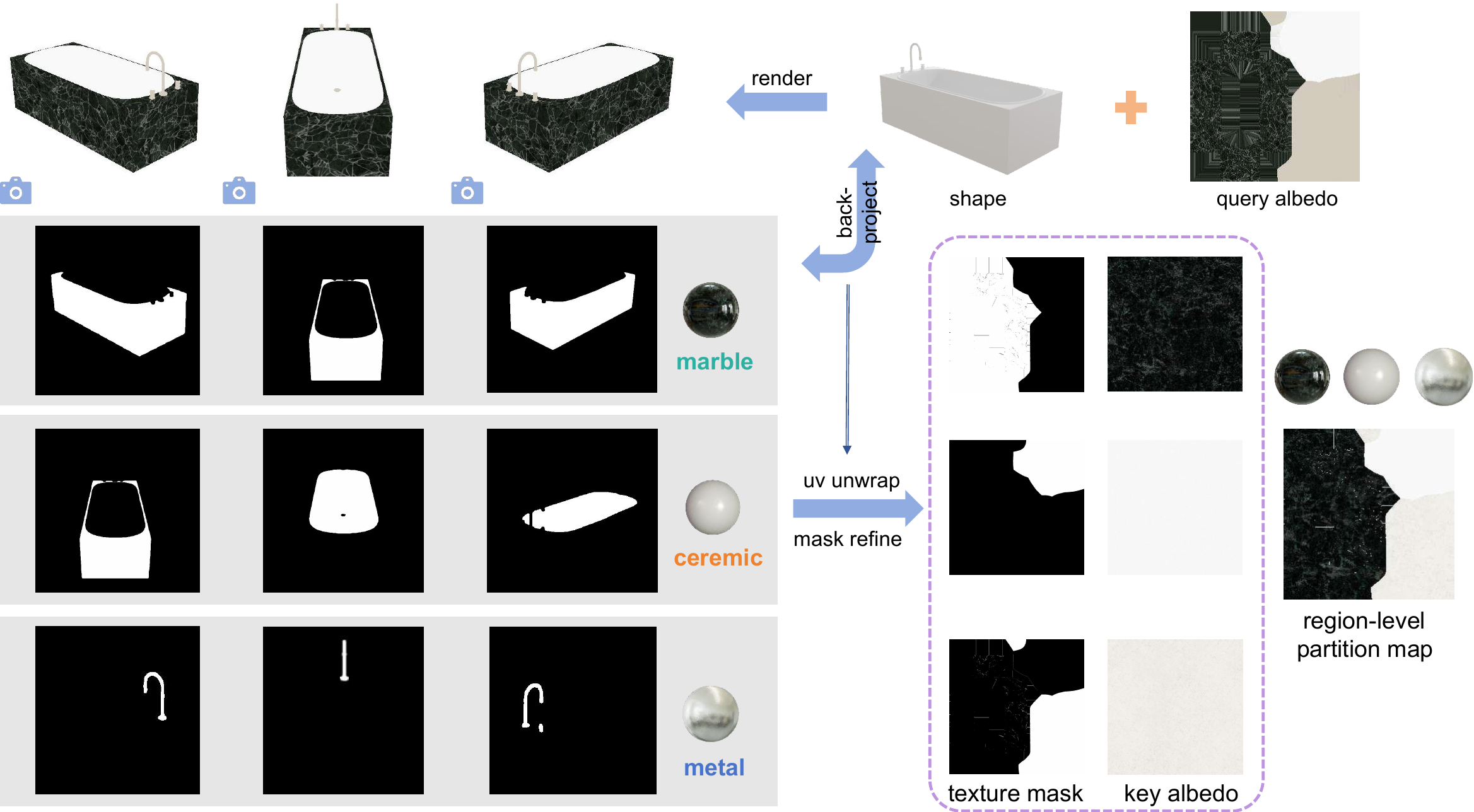}
  \vspace{-5pt}
   \caption{\textbf{Region-level texture partition module}. This module extracts and back-projects localized rendered images on to a 3D mesh, using UV unwrapping for texture segmentation, thereby resulting in precise partition map of different materials.}
   \vspace{-5pt}
   \label{fig:partition_module}
\end{figure*}

As illustrated in \Cref{fig:partition_module}, our process initiates with the rendering of a 3D object incorporating the original albedo (i.e. query albedo) from multiple perspectives. Following this rendering phase, we employ GPT-4V alongside a segmentor to derive segmented masks for the materials associated with each viewpoint. The subsequent step involves the extraction of regions masked in these images and their back-projection onto the mesh of the object. By examining the object from all acquired viewpoints and applying UV unwrapping techniques, we achieve preliminary segmentation of all materials. Subsequently, each material's segment is refined using a albedo-based mask refinement operation. Ultimately, by combining the segments of all materials, we obtain a region-level texture partition map, which serves to guide our subsequent, more detailed operations.

\begin{figure*}[!t]
  \centering
  \includegraphics[width=0.94\linewidth]{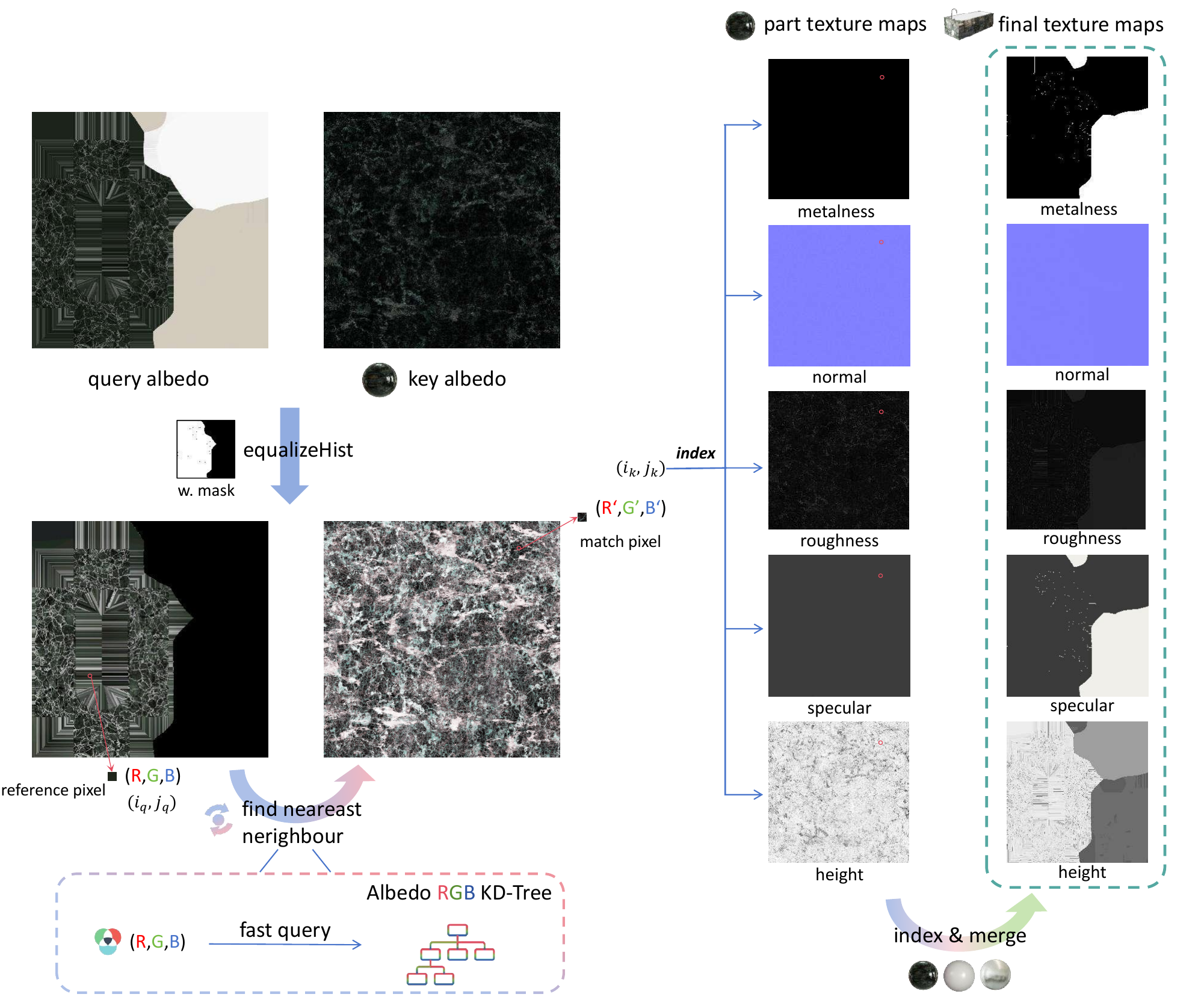}
  \vspace{-5pt}
   \caption{\textbf{Pixel-level albedo-referenced estimation module}. We generate spatially variant BRDF maps by referencing albedo maps, employing KD-Tree algorithm for efficient nearest neighbor searches, and normalizing colors via histogram equalization.}
   \vspace{-15pt}
   \label{fig:ref_module}
\end{figure*}

\subsection{Albedo-referenced Module Design}
\label{sup_b_2_albedo_refer}

As illustrated in \Cref{fig:ref_module}, we developed a pixel-level albedo-referenced estimation module, building upon the foundations laid out in \Cref{sup_b_1_texture_partition}. This module is inspired by a technique frequently employed by 3D artists, who often utilize albedo maps as a reference to generate images of other material properties. Accordingly, we designate the known albedo map as the query albedo, and the albedo corresponding to the region of interest in the material as the key albedo. 

The process to precisely obtain the final SVBRDF maps is divided into four steps: 1) Initially, to address potential gaps in color intensity between the two albedo maps, histogram equalization is employed to achieve a more uniform color distribution across the image. 2) Subsequently, for each pixel on the query albedo—termed a reference pixel—we seek the most similar neighboring color index on the key albedo. Given the high dimensionality of both maps (typically 1024x1024 pixels), a brute-force approach to this search would be computationally prohibitive. To this end, we accelerate the pixel query process using a KD-Tree algorithm, which organizes the RGB values of the key map's albedo into a KD-Tree for efficient nearest neighbor searches, reducing the computation time to under ten seconds. 3) The third step involves using the obtained indices to obtain corresponding values from the rest of the material maps. 4) Finally, by aggregating the query results for all material segments, we are able to generate the comprehensive spatially variant SVBRDF maps.

\subsection{Rendering Procedure Details}
\label{sup_b_m_setup}

In the context of computer graphics, ``albedo'' typically denotes the primary color of a material, a concept analogous to ``base color'' and ``diffuse'' in Physically-Based Rendering (PBR) paradigms, both representing the inherent color of the material under uniformly scattered illumination.

In accordance with workflow requirements, the inclusion of height maps, displacement maps, specular maps, and combinations of additional maps is optional. For objects demanding high surface details, such as the oil barrel in \Cref{fig:objaverse} and the stone horse in \Cref{fig:sup_more_tex}, displacement or height maps are incorporated and rendered using the mature 3D computer graphics software Blender engine. Regarding 2D-3D alignment techniques, including rasterization rendering, back-projection, and UV unwrapping detailed in \Cref{3_2_1:render_and_seg}, we adhere to the methodologies outlined in \cite{richardson2023texture}. Specifically, we utilize the Kaolin package (Kaolin \cite{jatavallabhula2019kaolin}) and back-projection with masks~\cite{richardson2023texture}, with the model learning rate set to 0.02 in \Cref{eq:texture_projection} and the difference coefficient $\sigma$ set to 0.1 in \Cref{eq:diff_cal}.

This approach facilitates the systematic generation of various material texture maps in the presence of only an albedo map, thereby ensuring consistency and realism within the Physically Based Rendering (PBR) workflow. Moreover, the flexible application of different map types and 2D-3D alignment techniques significantly enhances the detail and realism of rendered objects, effectively meeting the demands of diverse rendering scenarios.


\begin{figure*}[t]
  \centering
  \includegraphics[width=0.94\linewidth]{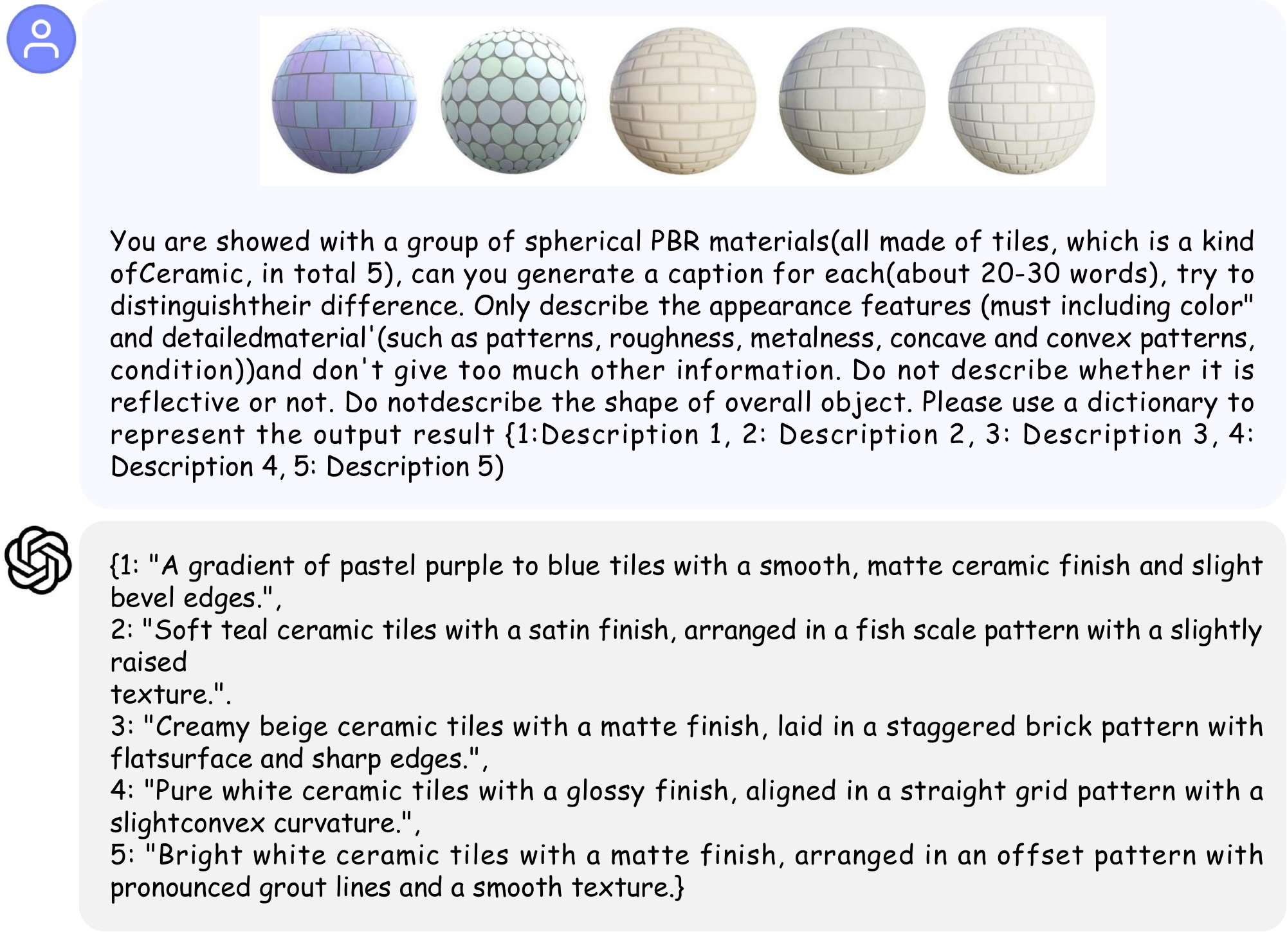}
  \vspace{-5pt}
   \caption{\textbf{Material captioning}. 
We demonstrate how to construct detailed material descriptions, registering material information in textual form, thus providing a convenient bridge when querying with multi-modal large language models.}
   \vspace{-15pt}
   \label{fig:gpt_material}
\end{figure*}

\subsection{Prompt Details for Multi-modal Large Language Models}
\label{sup_b_3_prompts_detail}

\begin{figure*}[htbp]
  \centering
  \includegraphics[width=0.94\linewidth]{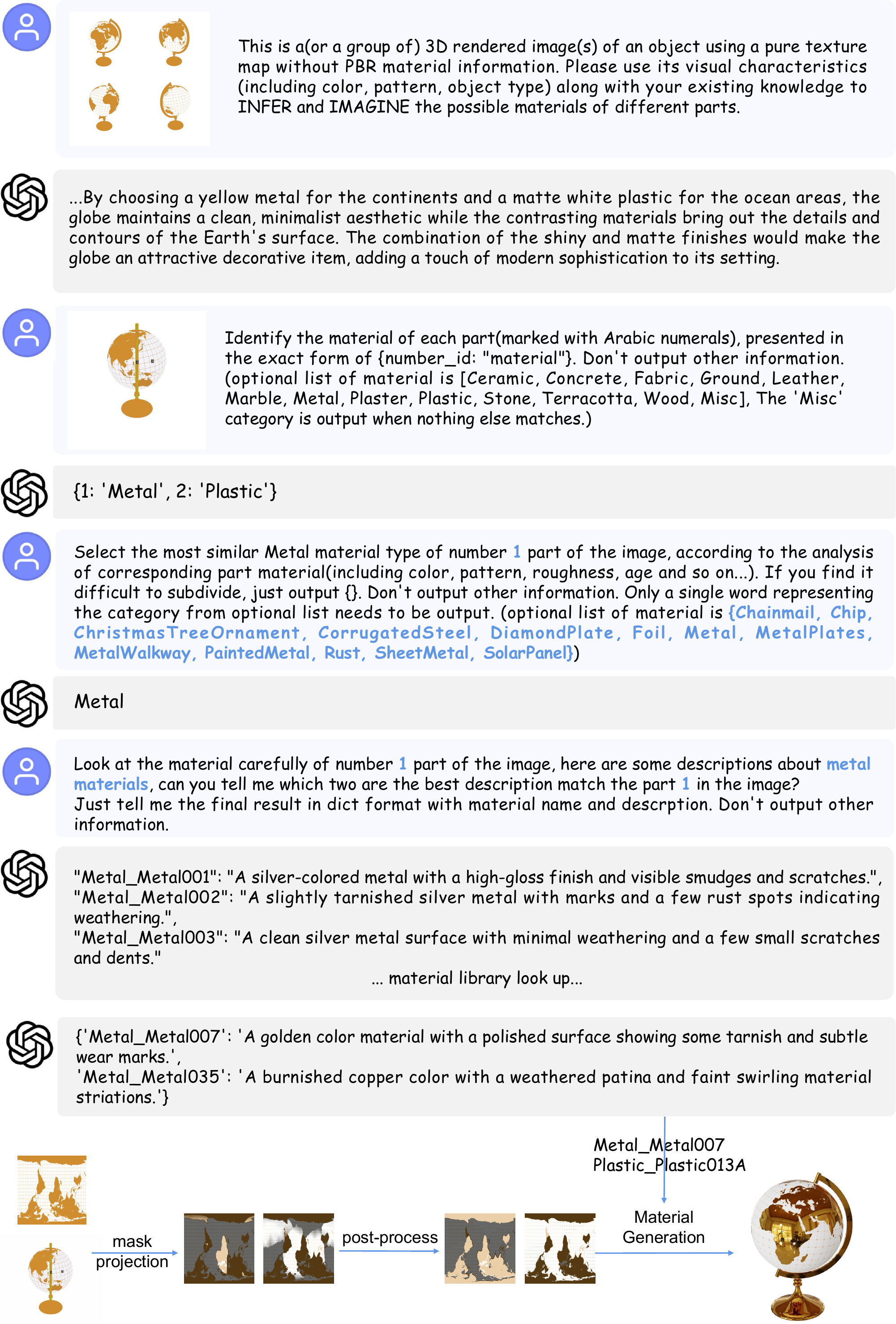}
  \vspace{-5pt}
   \caption{\textbf{Detailed prompts} of GPT-4V based material matching. Prompts in blue changes according to the current assigning part and GPT-4V's results.}
   \vspace{-15pt}
   \label{fig:gpt_full_process}
\end{figure*}

\noindent Prompts design has emerged as a pivotal factor for eliciting desired outcomes from MLLMs. This section delineates the intricacies of our prompt design for both material captioning and matching.

\noindent \textbf{Material captioning.} Due to limit context length for image tokens, it is currently impossible for MLLMs to directly memories thousands of images of material balls. To address this, we use GPT-4V to generate detailed captions for each material ball. As demonstrated in \Cref{fig:gpt_material}, we input concatenated images of material ball from same subcategory with prompt specifically tailored to highlight texture properties. This strategy guides GPT-4V to generate detailed caption for each material ball, distinguishing the subtle differences between them. These detailed captions are then registered into material library. 

\noindent \textbf{Material matching.} The MLLM-based material query process is exemplified through a simplified case, as illustrated in \Cref{fig:gpt_full_process}. Initially, GPT-4V is queried to identify the basic material for matching.  Following this preliminary matching, GPT-4V is queried for more specific materials using the names of subcategories as filters, which narrows down the selection to a few candidates within the same subcategory. For final material selection, GPT-4V is prompted with detailed captions from the library, directing it to allocate the most suitable material. This is accomplished in conjunction with a meticulously engineered segmentation process in UV space, ultimately facilitating MLLM-based material matching. It is noteworthy that prompting GPT-4V to navigate through a three-level tree structure has been proven to enhance both the efficiency and accuracy of the matching process, as opposed to directly selecting from thousands of materials directly.

\begin{figure*}[t]
  \centering
  \includegraphics[width=0.94\linewidth]{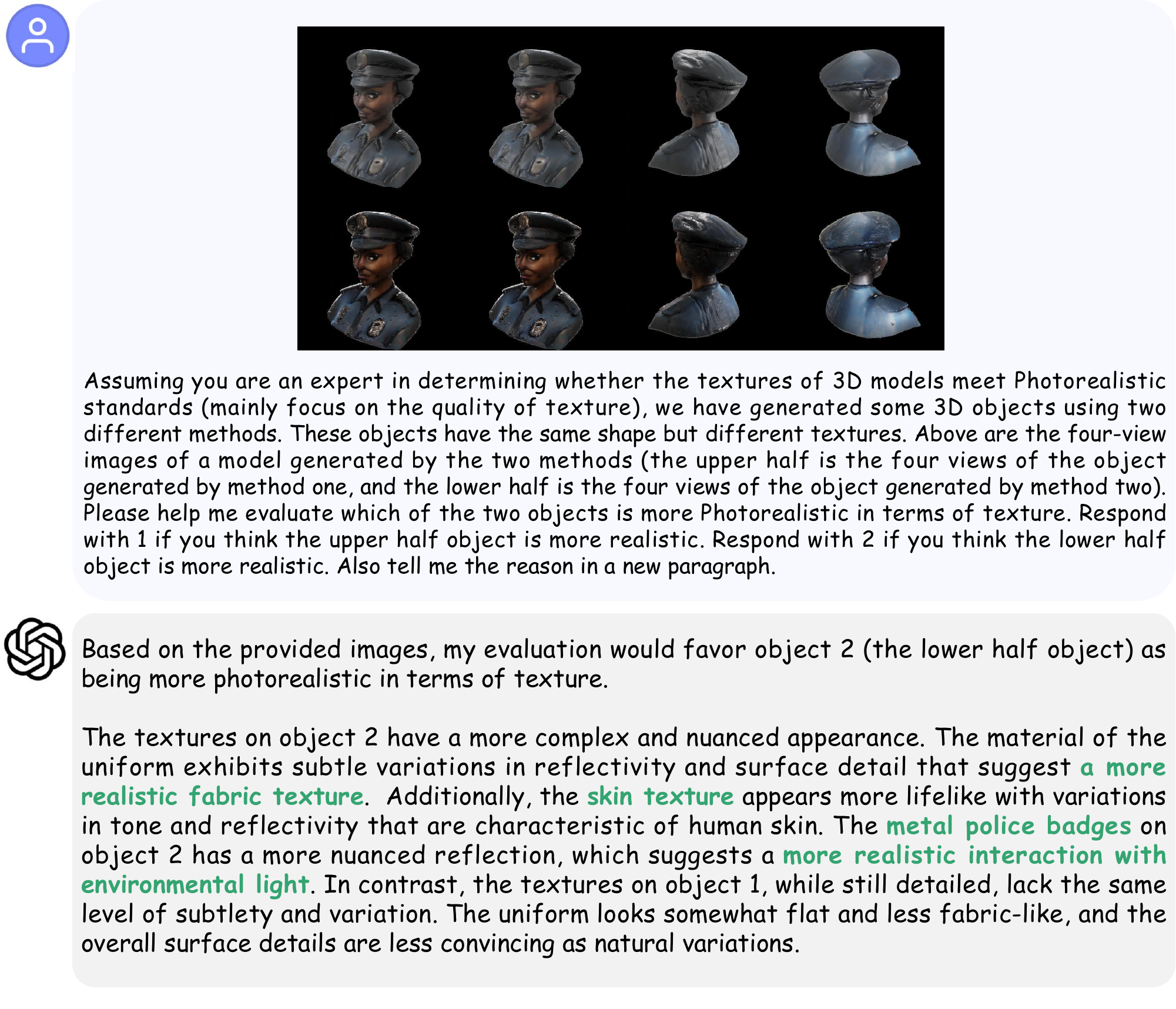}
  \vspace{-5pt}
   \caption{\textbf{GPT-4V based evaluation prompts.} We define a prompt for GPT-4V to generate human-aligned comparison over objects before and after our texture refinement.}
   \vspace{-15pt}
   \label{fig:gpt_eval_prompt}
\end{figure*}

\subsection{Evaluation Details of Quantitative Results}
\label{sup_b_4_evaluation_detail}

\label{sup:gpt_evaluation}
\noindent \textbf{Evaluation details.} For the purpose of evaluating objects pre- and post-refinement, we render four view images of each object and concatenate them vertically to facilitate a comprehensive assessment. We craft a specific prompt to guide GPT-4V in conducting an impartial comparison of the texture quality between two objects. As demonstrated in \Cref{fig:gpt_eval_prompt}. GPT-4V's advanced capabilities enable it to differentiate between the two objects, providing comparison results that closely align with assessments made by human experts. In the scenario evaluated, GPT-4V successfully identifies all enhancements (metal badges, human skin, and fabric textures).

\section{Additional Related Works}

In the context of extracting real-world materials from single images, the methodologies most closely related to our work are Material Palette \cite{lopes2023material} and Photoshape \cite{park2018photoshape}. Material Palette enables the extraction of materials at the region level from a single image, generating tilable texture maps for corresponding areas. Photoshape, on the other hand, automates the assignment of real materials to different parts of a 3D shape by training a material classification network. However, this approach needs the training of a material classifier, which is constrained to a limited set of object categories (e.g., chairs) and material types. Besides, both methods require rendered images of objects with real materials  as input. 

\begin{figure*}[t]
  \centering
  \includegraphics[width=\linewidth]{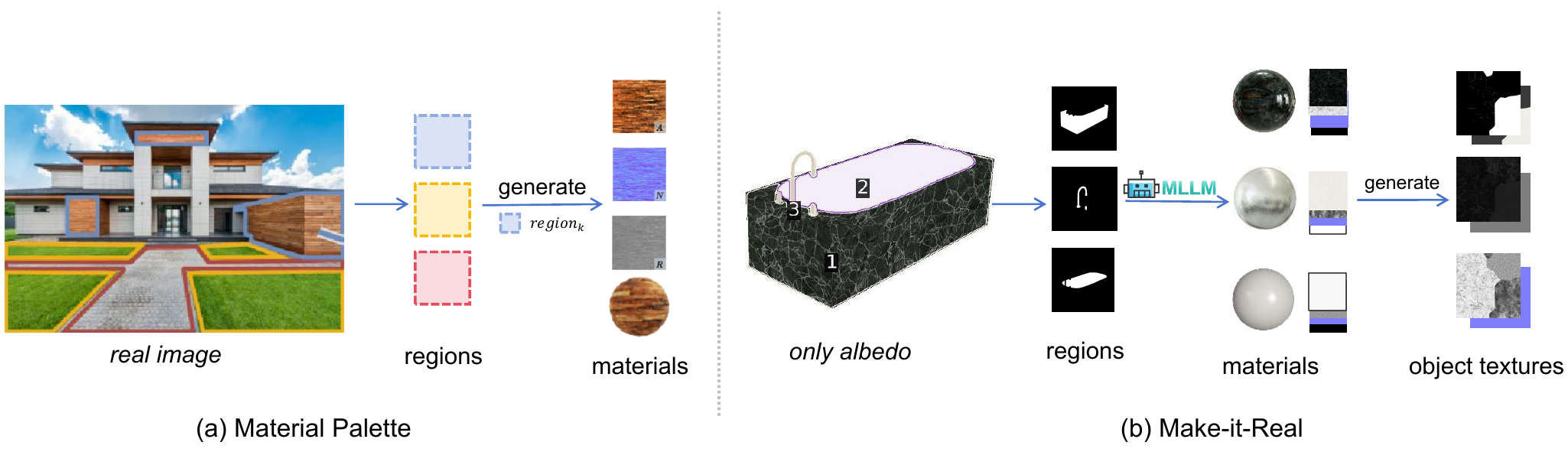}
  \vspace{0pt}
   \caption{\textbf{Comparison between previous method and \methodname.} We demonstrate the distinctions between Material Palette\cite{lopes2023material} and our method in terms of material identification and extraction. Our overall pipeline presents a more challenging task, where the input is a rendered image with only albedo information, and the output consists of textures for the entire object.}
   \vspace{-8pt}
   \label{fig:comp}
\end{figure*}

Our problem setting poses a more challenging task, as illustrated in \Cref{fig:comp}, which involves identifying and recovering the original material properties from images of objects that only have a albedo map, in addition to generating different material maps for the entire object. Humans have the capacity to intuitively infer the underlying material properties of 3D objects from single images containing only shape and basic colors. This capability is attributed to our robust material recognition abilities and comprehensive prior knowledge of object categories, colors, and material attributes. Building on this concept, our approach harnesses the potent image recognition capabilities and prior knowledge inherent in large-scale multimodal language models to efficiently execute region-level material extraction and identification, which is further enhanced by subsequent 2D-3D alignment and albedo-referenced techniques to generate and apply material texture maps for physically realistic rendering of 3D objects. Furthermore, ~\cite{lopes2023material} extracts the material takes 3 to 4 minutes, and process an image containing three materials takes more than 10 minutes. In contrast, our method takes only about 1 to 2 minutes to match and generate all the materials.

\section{Additional Experiments}

\subsection{Effects of Different Texture Maps}
\label{sup_d_1}

\begin{figure*}[t]
  \centering
  \includegraphics[width=0.98\linewidth]{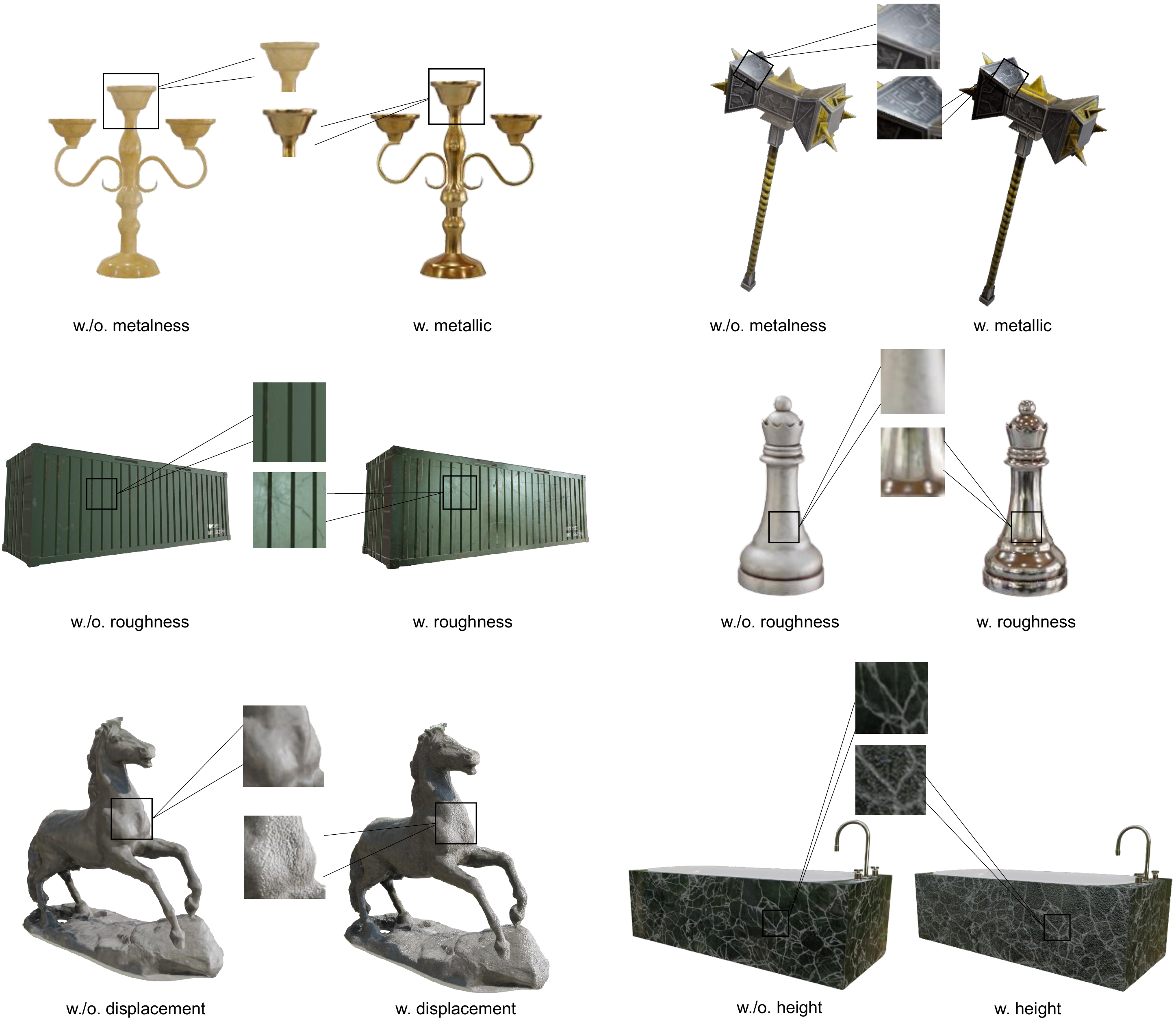}
  \vspace{0pt}
   \caption{\textbf{Effects of different texture maps.} We evaluate the effects of metalness, roughness, and displacement/height maps on the appearance of 3D objects.}
   \vspace{-16pt}
   \label{fig:sup_more_tex}
\end{figure*}

We validate the impact of various texture maps generated by \methodname on the appearance of 3D objects, as illustrated in \Cref{fig:sup_more_tex}. For each example, the first column lacks the corresponding map enhanced by \methodname, while the second column includes the same conditions with the specific material map. Initially, we compare the effect of metalness on object appearance. We observe that the objects on the right, with a higher metalness map value, exhibit higher reflectivity, and the reflected light's color is similar to the material's own color, closely resembling real-world objects and appearing more aesthetically pleasing. In contrast, the objects on the left without the metalness map have lower reflectivity, and the reflected light tends to be white. 

Next, we examine the impact of the roughness map, which controls the smoothness of the material's surface. We observe that the silver chess piece on the right, with a low roughness texture map, becomes smooth, and together with metalness, produces a mirror-like reflection effect. On the other hand, the box on the left with a roughness map exhibits changes in light dispersion on its surface, with some areas showing highlights, while also adding and enriching scratch texture details on the surface. 

Furthermore, we compare the effects of displacement and height on objects, both of which are usually optional and can also impact object appearance. Height maps typically use grayscale values to represent the surface's relative height, simulating a relief effect through changes in lighting and shadows. As shown in the last row on the right, the object with a height map has an uneven surface, enhancing the sense of depth. Displacement mapping is a more powerful technique that changes the vertex positions of the geometry based on the map's values, creating realistic relief effects. As shown on the left, the stone sculpture with displacement mapping exhibits very realistic details and a sense of relief.


\section{More Qualitive Results}
\label{sup:more_results}

\subsection{\methodname for Existing 3D Assets}
\begin{figure*}[htbp]
  \centering
  \includegraphics[width=0.98\linewidth]{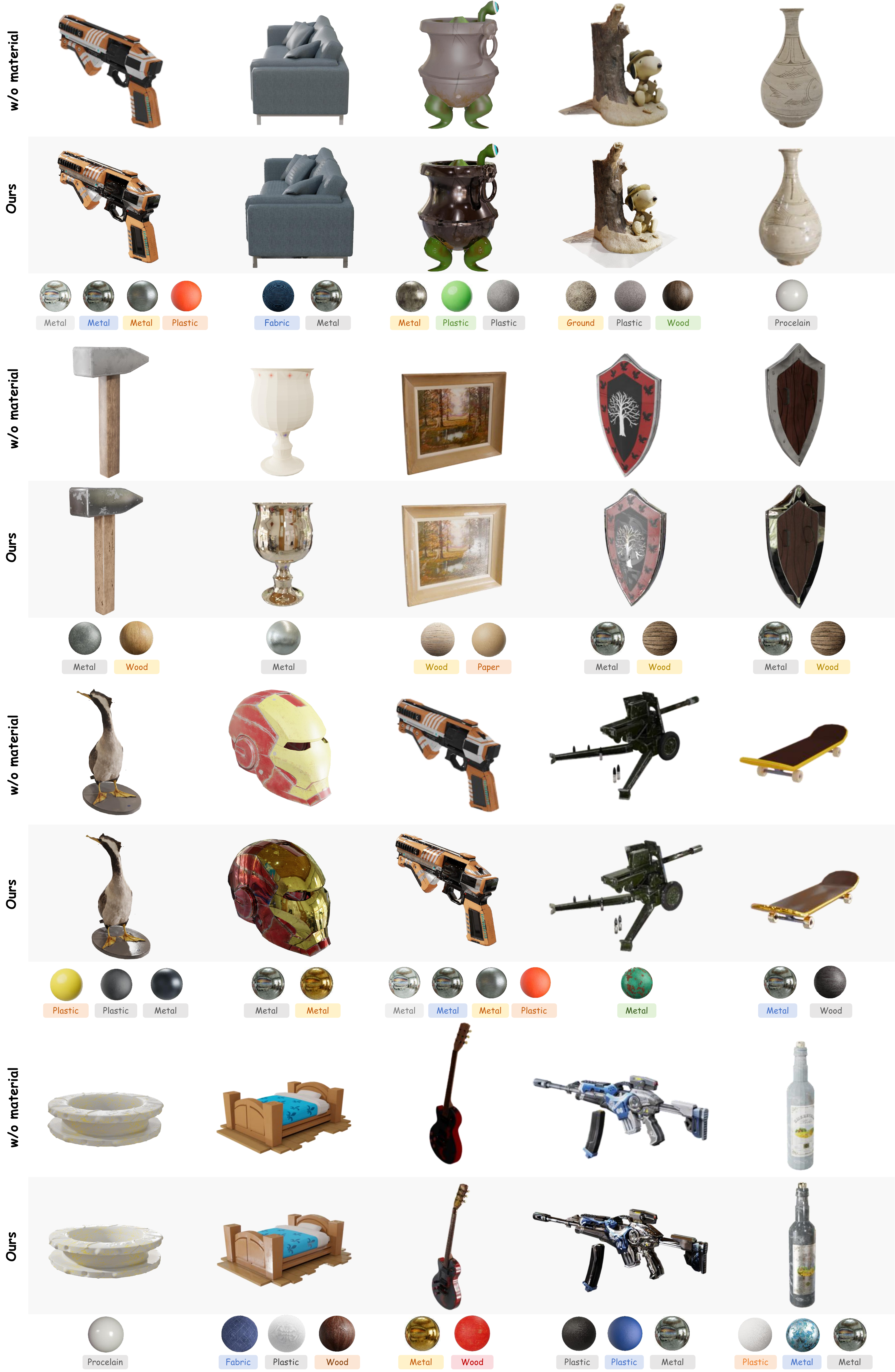}
  \vspace{0pt}
   \caption{\textbf{More qualitative results of \methodname refining existing 3D assets without material.} Objects are selected from Objaverse\cite{objaverse} with albedo only.}
   \vspace{-8pt}
   \label{fig:sup_case_obja1}
\end{figure*}

\begin{figure*}[htbp]
  \centering
  \includegraphics[width=0.98\linewidth]{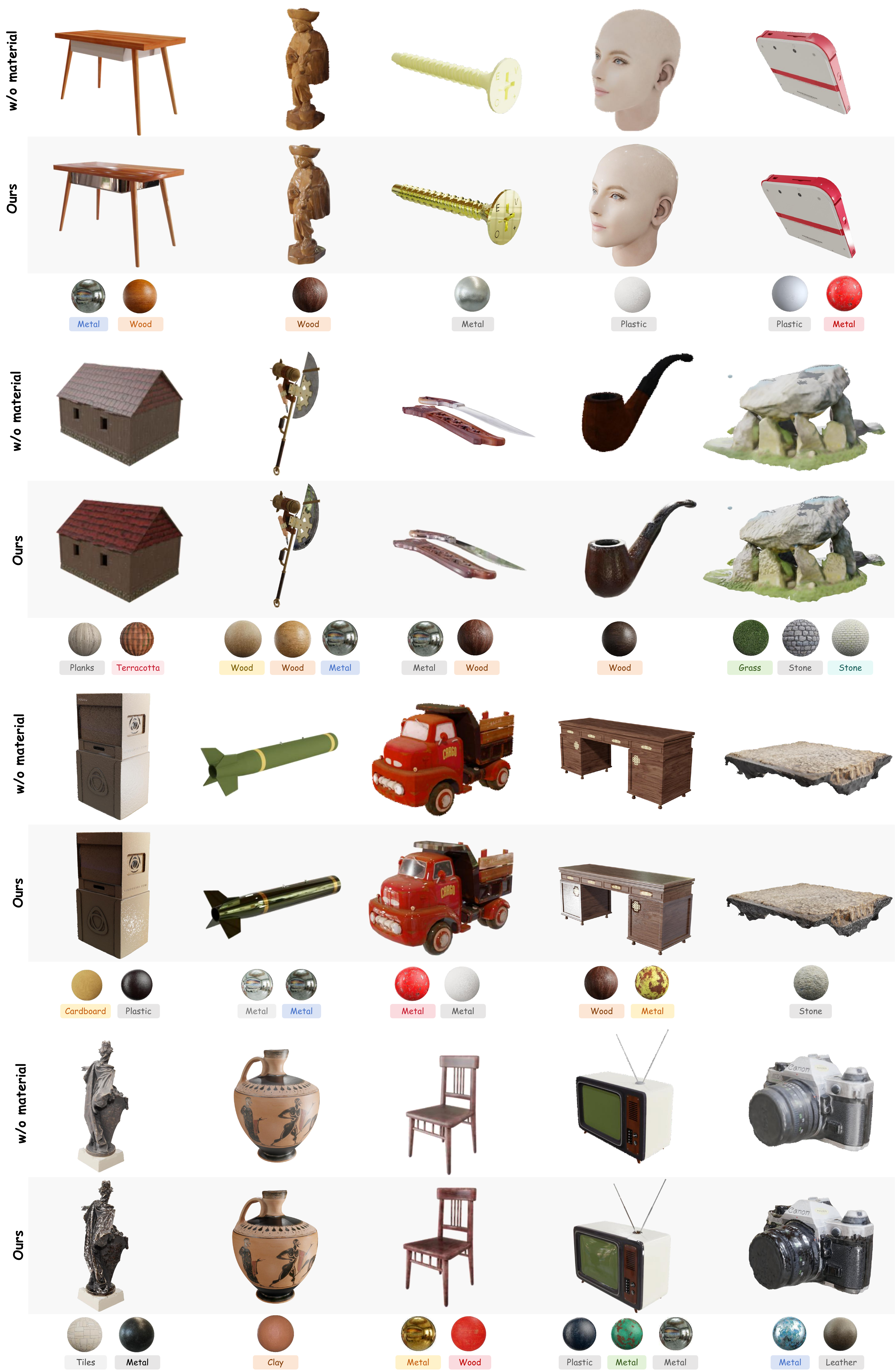}
  \vspace{0pt}
   \caption{\textbf{More qualitative results of \methodname refining existing 3D assets without material.} Objects are selected from Objaverse\cite{objaverse} with albedo only.}
   \vspace{-8pt}
   \label{fig:sup_case_obja2}
\end{figure*}

In this section, we show more qualitative results of \methodname for existing 3D assets from \cite{objaverse}. Results are shown in \Cref{fig:sup_case_obja1,fig:sup_case_obja2}. The first row presents the original 3D object with only a albedo map, while the second row showcases the object enhanced by our method with various material maps.

\begin{figure*}[htbp]
  \centering
  \includegraphics[width=0.98\linewidth]{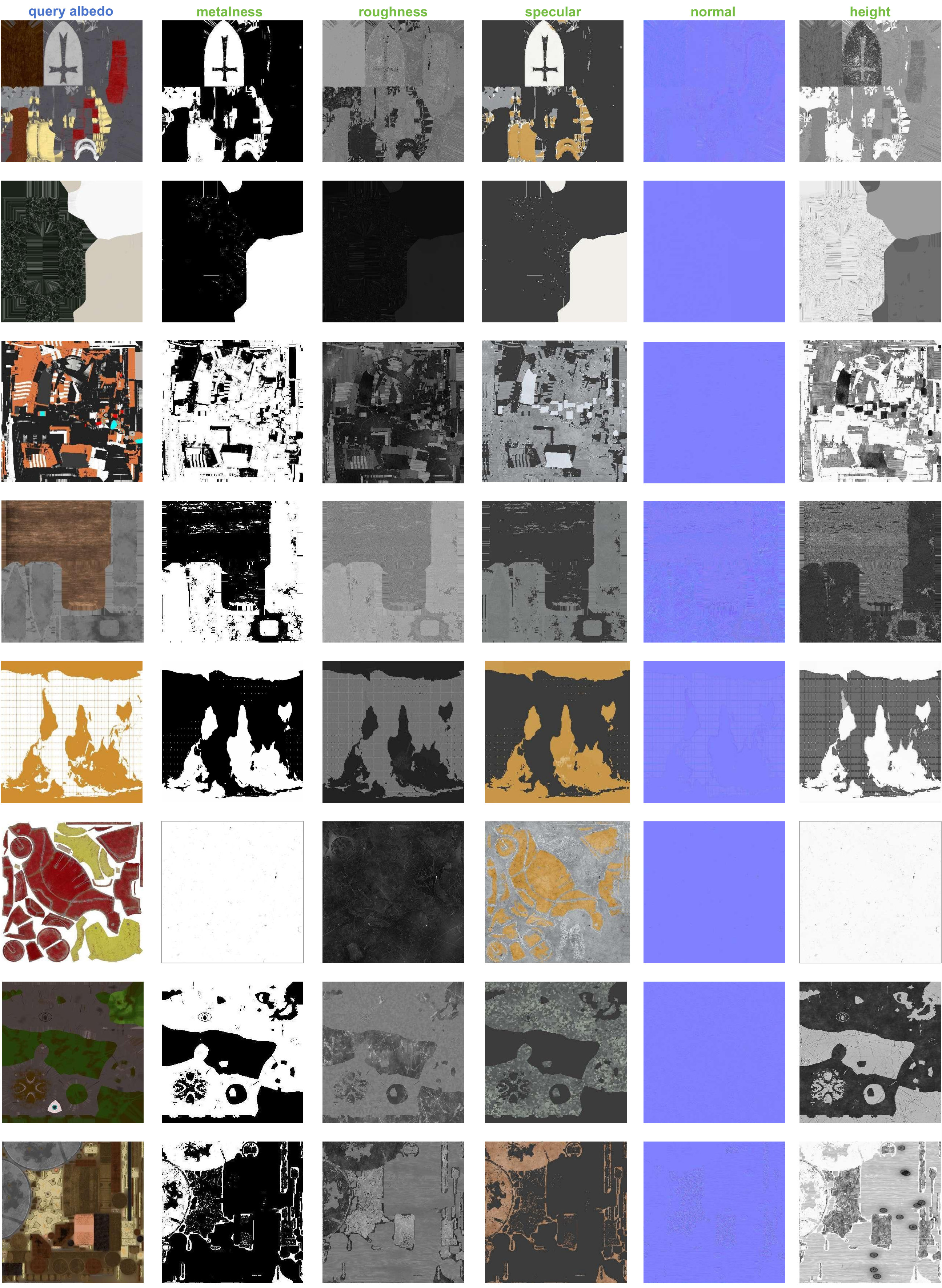}
  \vspace{0pt}
   \caption{\textbf{Visualization of generated texture maps.} The first column represents the original query albedo map of 3D objects, while the subsequent columns showcase the corresponding material maps generated by \methodname.}
   \vspace{-8pt}
   \label{fig:sup_visualize1}
\end{figure*}

\begin{figure*}[htbp]
  \centering
  \includegraphics[width=0.98\linewidth]{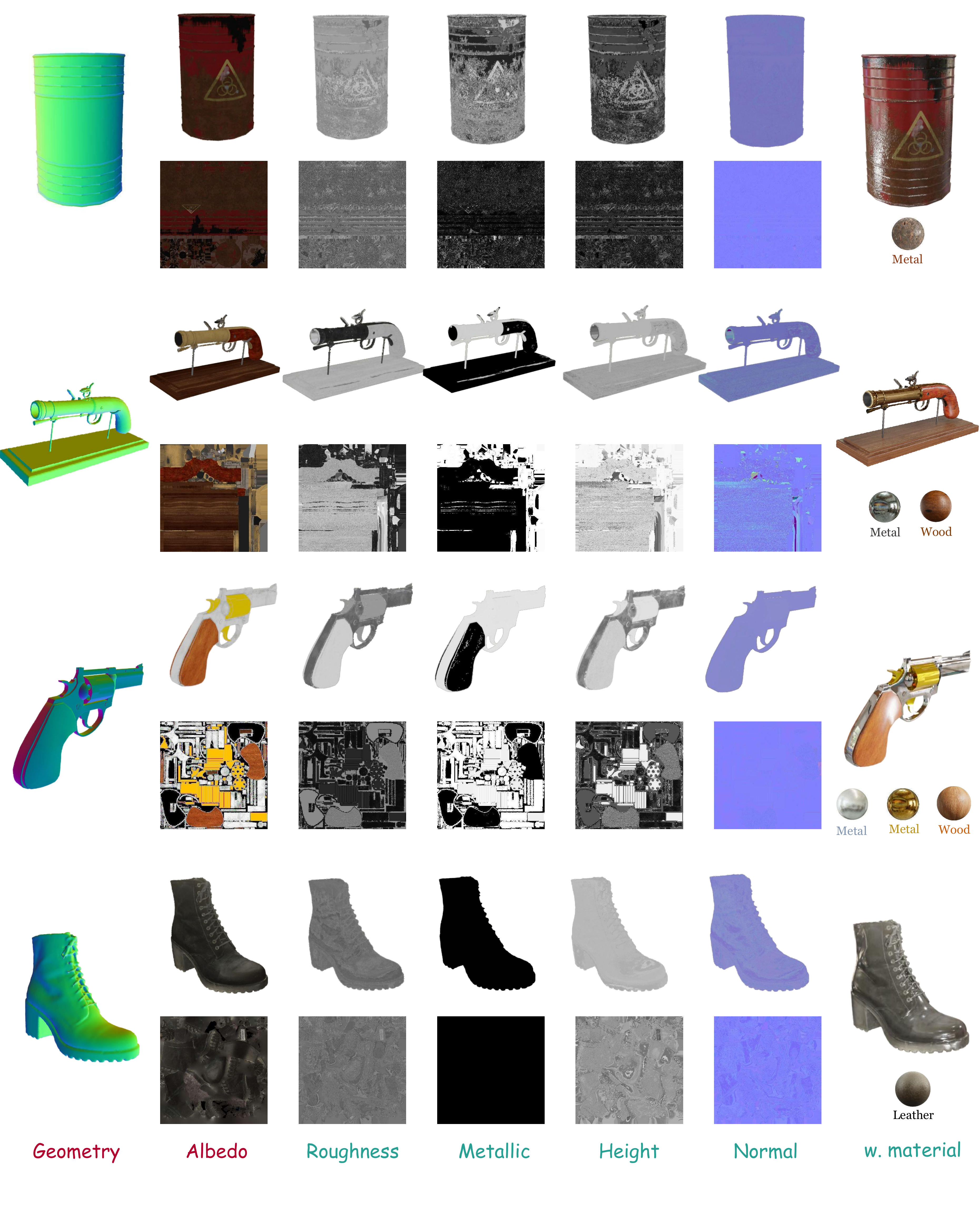}
  \vspace{0pt}
   \caption{\textbf{Visualization of generated texture maps.} We show part of the SVBRDF material maps generated by \methodname and the final rendering results. We displayed the texture maps and the corresponding 3D rendering effects. The albedo is the input, and the following four columns show the material effects in the UV space and on the 3D object.}
   \vspace{-8pt}
   \label{fig:sup_visualize2}
\end{figure*}

\subsection{Visualization of generated texture maps.} 
In this section, we show visualization of generated texture maps in \Cref{fig:sup_visualize1,fig:sup_visualize2}. Our approach, guided by the query albedo reference, produces material maps with distinct material partitions and maintains a distribution consistent with the albedo map. As a result, the enhanced object exhibits both realism and texture consistency.

\subsection{Limitations and future work.}
\label{sec:limitation}
While promising results have been achieved by MLLMs for texture assignment through \methodname. Our work still faces several unresolved challenges. First, although our method can achieve appropriate texture assignment for albedo-only model, it does not support reverse transform from shaded texture map to albedo map for generated 3D objects. This causes problem of assigning different materials to the dark shadow and highlight area when generated object with mesh already shaded in different lighting conditions. Second, we find base model quality is essential for MLLM to assign correct materials. When base 3D object is of low quality, it is difficult for MLLMs to identify object properties when ground truth text prompt describing the object is not available. Additionally, there has been limited progress in material segmentation in 2D and 3D demoain\cite{sharma2023materialistic, li2024materialseg3d}. The material segmentation algorithm included in our proposed pipeline can serve as a useful baseline and provide inspiration for future work. We also explore the integration of the pipeline with 3D segmentation networks, which shows promising results. For future work, we consider methods mitigating these challenges as well as adding user friendly control into our fully-automatic pipeline.

\clearpage
\newpage
\section{Broader Impacts}
\label{sup:broader_impacts}
\textbf{Potential positive societal impacts:} The proposed method facilitates more realistic and accurate representations of materials in 3D models, benefiting industries such as gaming, virtual reality, and film, leading to more immersive and engaging experiences. By automating the material assignment process, Make-it-Real significantly reduces the time and effort required for 3D content creators, allowing for more efficient workflows and enabling creators to focus on more creative aspects of their work. This approach can democratize high-quality 3D content creation by making advanced material application techniques accessible to a broader range of users, including those without specialized skills in graphic software.

\textbf{Potential negative societal impacts:} The improved realism in 3D assets could be exploited for creating highly convincing fake visuals or deepfakes, which might be used in disinformation campaigns or to mislead audiences.  There is a risk that the materials generated could inadvertently perpetuate biases or stereotypes if the training data for GPT-4V includes biased representations of certain materials or objects. As the method involves processing and recognizing visual data, there could be concerns regarding the privacy of any real-world images used as inputs, particularly if they contain sensitive or personal information.

\end{document}